\documentclass[letterpaper,journal]{IEEEtran}
\usepackage{booktabs}
\usepackage{amsmath,amsfonts}
\usepackage{algorithmic}
\usepackage{algorithm}
\usepackage{array}
\usepackage{multirow}
\usepackage[caption=false,font=normalsize,labelfont=sf,textfont=sf]{subfig}
\usepackage{textcomp}
\usepackage{stfloats}
\usepackage{url}
\usepackage{verbatim}
\usepackage{graphicx}
\usepackage{cite}

\usepackage[colorlinks=true, citecolor=blue, linkcolor=blue, urlcolor=blue]{hyperref}
\begin{document}

\title{BELIEF: Structured Evidence Modeling and Uncertainty-Aware Fusion for Biomedical Question Answering}

\author{
Chang Zong~\href{https://orcid.org/0000-0001-7757-0659}{\includegraphics[height=2.5mm]{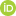}},~\IEEEmembership{Member,~IEEE,}
Hao Ning~\href{https://orcid.org/0009-0001-2041-1267}{\includegraphics[height=2.5mm]{Fig_orcid_icon.png}},~\IEEEmembership{Graduate Student Member,~IEEE,}
Siliang Tang~\href{https://orcid.org/0000-0002-7356-9711}{\includegraphics[height=2.5mm]{Fig_orcid_icon.png}}\\
Jie Huang~\href{https://orcid.org/0000-0001-9717-8355}{\includegraphics[height=2.5mm]{Fig_orcid_icon.png}},~\IEEEmembership{Member,~IEEE,}
and Jian Wan~\href{https://orcid.org/0000-0001-9882-3029}{\includegraphics[height=2.5mm]{Fig_orcid_icon.png}}
\thanks{
Chang Zong, Hao Ning, and Jie Huang are with the School of Computer Science and Technology, Zhejiang University of Science and Technology, Hangzhou 310023, China.
}
\thanks{
Siliang Tang is with the College of Artificial Intelligence, Zhejiang University, Hangzhou 310027, China.
}
\thanks{
Jian Wan is with the Zhejiang Key Laboratory of Biomedical Intelligent Computing Technology, Zhejiang University of Science and Technology, Hangzhou 310023, China.
}
\thanks{
Corresponding author: Hao Ning (e-mail: ninghao@zust.edu.cn).
}
}

\maketitle
\begin{abstract}
Biomedical question answering often requires decisions from retrieved literature whose relevance, quality, and support for candidate answers are uneven.
Most retrieval-augmented large language model (LLM) methods feed this literature to the model as flat text, leaving evidence reliability and remaining uncertainty largely implicit. We propose BELIEF, a structured evidence modeling and uncertainty-aware fusion framework for closed-set biomedical question answering. Rather than treating retrieved documents as undifferentiated context, BELIEF converts them into evidence objects that record clinical attributes, source quality, question relevance, support strength, and the associated candidate hypothesis. These evidence objects provide a shared basis for two complementary reasoning paths. The symbolic path constructs reliability-weighted basic probability assignments based on Dempster--Shafer (D-S) theory over a finite answer space and performs uncertainty-aware symbolic evidence fusion to estimate belief and residual uncertainty. The neural path uses the same structured evidence for LLM-based semantic inference, while a reliability-aware arbitration module reconciles the symbolic and neural outputs according to belief strength, uncertainty, evidence reliability, and semantic consistency.
Experiments on PubMedQA, MedQA, and MedMCQA with five general-purpose LLM backbones show that BELIEF obtains the best result in 25 of 30 backbone--dataset--metric settings.
Comparisons with biomedical-domain models indicate that BELIEF is competitive on MedQA and MedMCQA, while specialized biomedical pretraining remains advantageous on PubMedQA.
Ablation, complementarity, uncertainty-stratified, and cost analyses further show that BELIEF improves retrieved-evidence utilization by making evidence structure, path disagreement, and decision uncertainty explicit.
\end{abstract}


\section{Introduction}

\IEEEPARstart{R}ECENT advances in large language models (LLMs) have substantially improved question answering (QA), especially when combined with retrieval-augmented generation (RAG) techniques~\cite{brown2020language,touvron2023llama,lewis2020retrieval}. 
By grounding responses in external biomedical evidence, RAG-based methods can improve factual consistency~\cite{gao2024retrieval,asai2023self}. 
However, most existing LLM-based retrieval and reasoning methods still treat retrieved evidence as flat textual context and rely on implicit neural aggregation, making it difficult to explicitly model evidence reliability, residual uncertainty, and multi-evidence support for candidate decisions.

This limitation is particularly important in biomedical QA. 
From the perspective of evidence-based medicine (EBM)~\cite{sackett1996evidence,guyatt2011grade}, biomedical decision-making requires assessing evidence quality, relevance, study characteristics, and uncertainty rather than only retrieving topically related documents. 
When retrieved evidence is incomplete, heterogeneous, or weakly grounded, flat-context aggregation may lead to unstable or overconfident predictions~\cite{kim2025medical,huang2025survey,xu2025large}. 
Therefore, biomedical QA can be reformulated as a structured evidence to-decision problem, where unstructured literature is transformed into comparable evidence objects, reliability and support are quantified, and uncertainty-aware decision signals are exposed.

\begin{figure}[t]
\centering
\includegraphics[width=1.0\linewidth]{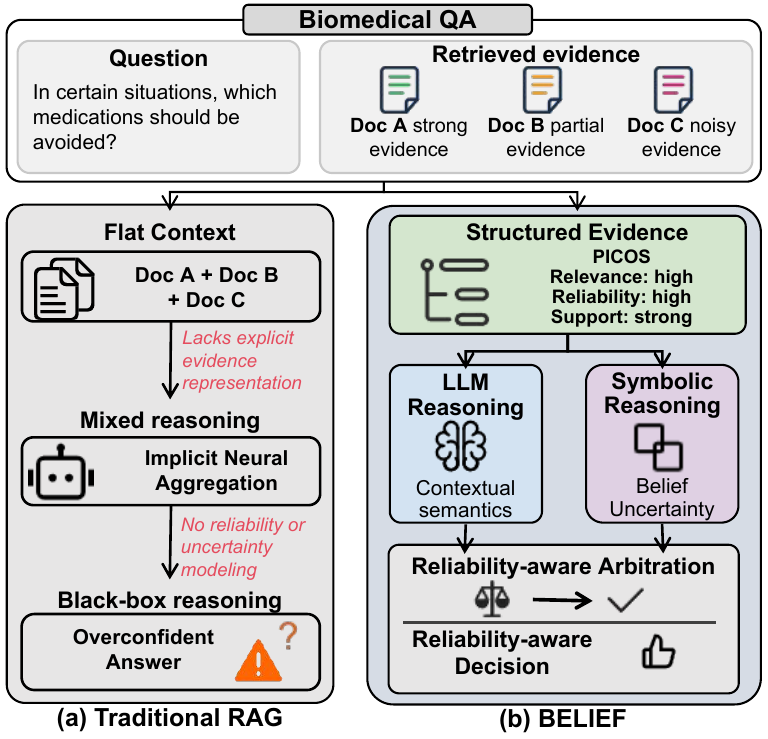}
\caption{Traditional RAG treats retrieved evidence as flat context, whereas BELIEF structures evidence and exposes reliability and uncertainty signals.}
\label{fig:evi_overview}
\end{figure}

A promising direction is to integrate symbolic evidence reasoning with neural semantic reasoning in a unified framework.~\cite{shi2025final,xiao2022generalized} 
Symbolic reasoning provides explicit mechanisms for evidence aggregation, reliability estimation, and uncertainty modeling, whereas neural LLM-based reasoning captures semantic relationships and contextual biomedical knowledge. 
Dempster--Shafer (D-S) theory~\cite{dempster2008upper,shafer1976mathematical} is well aligned with this goal because it supports reliability-aware evidence combination and represents both belief over hypotheses and residual uncertainty. 
However, directly coupling D-S fusion with LLM-based reasoning remains non-trivial, since retrieved biomedical evidence must first be converted into computable mass assignments and the symbolic and neural paths produce different types of reasoning signals.

This motivates the central research question of this work: how can principled symbolic evidence reasoning be integrated with the flexible semantic reasoning capability of LLMs to support evidence-grounded and uncertainty-aware biomedical QA?
To answer this question, we reformulate biomedical QA as an evidence-driven decision-making problem under uncertainty, where structured evidence aggregation and neural semantic reasoning are jointly modeled and coordinated.
This reformulation introduces two key technical challenges.

\textit{Challenge I: Structured Evidence Representation for Neuro-Symbolic Reasoning.}
Retrieved biomedical literature is unstructured and cannot be directly used for quantitative evidence fusion. 
It must therefore be transformed into structured, comparable, and computable evidence representations that preserve clinically meaningful attributes such as study design, relevance, support strength, and hypothesis association~\cite{su2024knowledge}.

\textit{Challenge II: Reliable Integration of Complementary Reasoning Signals.}
Symbolic D-S aggregation and neural LLM reasoning may produce different predictions because they rely on different evidence representations and reasoning mechanisms.
The final decision therefore needs an integration mechanism that can judge when to trust symbolic belief signals and when to rely more on neural semantic inference.

To address these challenges, we propose BELIEF (Biomedical Evidence Modeling with Uncertainty-Aware Evidence Fusion), a structured evidence modeling and uncertainty-aware fusion framework for evidence-grounded biomedical QA, as illustrated in Fig.~\ref{fig:evi_overview}.
BELIEF explicitly integrates two complementary reasoning paths.
First, it constructs structured evidence representations guided by EBM principles, including PICO-based query abstraction and evidence-level attributes such as study design and relevance, thereby transforming heterogeneous biomedical literature into computable evidence objects.
Second, based on the structured evidence, the symbolic D-S path performs reliability-weighted evidence fusion to estimate belief and residual uncertainty, while the neural LLM path conducts semantic inference over the same evidence.
Finally, a reliability-aware arbitration module integrates symbolic and neural reasoning outputs by considering symbolic belief, residual uncertainty, and semantic consistency.

Experiments and diagnostic analyses on PubMedQA, MedQA, and MedMCQA show that BELIEF improves retrieved-evidence utilization across multiple backbones and yields larger gains over Naive RAG in higher-uncertainty groups.

In summary, the main contributions of this work are fourfold:

\begin{itemize}

\item We formulate closed-set biomedical QA as a structured evidence-to-decision problem under uncertainty, where heterogeneous retrieved literature is represented as reliability-weighted evidence over a finite answer space, making evidence reliability and residual uncertainty explicit.

\item We develop a structured evidence modeling scheme that transforms unstructured biomedical literature into evidence objects with clinical attributes, source quality, relevance, support strength, and hypothesis association, providing a shared computable basis for both symbolic D-S fusion and neural semantic reasoning.

\item We propose an uncertainty-aware fusion mechanism that derives D-S basic probability assignments from structured evidence, estimates belief and residual uncertainty, and integrates them with LLM-based semantic reasoning through reliability-aware arbitration, enabling more robust use of complementary reasoning signals.

\item We conduct experiments on PubMedQA, MedQA, and MedMCQA across multiple LLM backbones, with ablation and diagnostic analyses showing that BELIEF consistently improves retrieved-evidence utilization and delivers strong performance under heterogeneous and uncertain evidence conditions.

\end{itemize}

\section{Related Work}

\subsection{Biomedical Question Answering and Retrieval-Augmented LLMs}

Biomedical question answering has become an important benchmark for evaluating the knowledge and reasoning capability of language models in medical domains~\cite{jin2022biomedical}.
Datasets such as PubMedQA~\cite{jin2019pubmedqa}, MedQA~\cite{jin2021disease}, and MedMCQA~\cite{pal2022medmcqa} cover literature-grounded yes/no/maybe decisions, professional medical examination questions, and large-scale multiple-choice clinical reasoning.
Recent biomedical LLMs, including BioMistral~\cite{labrak2024biomistral}, MEDITRON~\cite{chen2023meditron}, and Med-PaLM~\cite{singhal2025toward}, improve domain-specific knowledge through pretraining, instruction tuning, or alignment.
Retrieval-augmented generation (RAG) further enhances factual grounding by incorporating external evidence into generation~\cite{lewis2020retrieval,gao2024retrieval,fan2024survey}.
Variants such as Self-RAG~\cite{asai2023self}, RAT~\cite{wang2024rat}, and CRAG~\cite{yan2024corrective} introduce reflection, iterative reasoning, or correction mechanisms. Recent extensions further optimize LLM reasoning through multi-path optimization or global planning strategies that decompose questions into atomic retrieval tasks~\cite{liao2026enhancing, li2025framework}.
However, these methods typically use retrieved evidence as textual context and do not explicitly represent evidence reliability or residual uncertainty over candidate hypotheses.

\subsection{Structured Evidence Modeling and Evidence-Based Medicine}

Evidence-based medicine emphasizes that clinical decisions should consider study design, evidence quality, population relevance, and outcome consistency in addition to textual evidence content~\cite{sackett1996evidence,guyatt2011grade,page2021prisma}.
Structured representations such as PICO and PICOS organize clinical questions and evidence according to population, intervention, comparison, outcome, and study design~\cite{richardson1995well,nye2018corpus,huang2006evaluation}.
Prior NLP and biomedical informatics studies have shown that PICO-style structures, trial characteristics, risk-of-bias signals, and key findings can be extracted from medical literature for evidence synthesis~\cite{nye2018corpus,lehman2019inferring,marshall2016robotreviewer,marshall2020trialstreamer}.Case-based reasoning (CBR) frameworks also support clinical transparency by retrieving and adapting similar historical patient cases to justify recommendations~\cite{pradeep2025empowering}.
These studies demonstrate the feasibility of converting unstructured biomedical literature into structured evidence representations.
However, most LLM-based QA systems still use retrieved evidence primarily as flat textual context and do not convert evidence quality, relevance, support strength, and hypothesis association into computable signals for downstream reasoning.
BELIEF addresses this gap by treating retrieved documents as structured evidence objects whose clinically meaningful attributes are explicitly modeled before uncertainty-aware fusion and arbitration.

\subsection{Uncertainty-Aware Evidence Fusion and Neuro-Symbolic Reasoning}

Uncertainty modeling is essential for decision making under heterogeneous evidence with varying reliability and incomplete support.
D-S theory has been widely used for evidence combination and multi-source uncertainty modeling~\cite{shafer1976mathematical,sentz2002combination,li2024inconsistency}, but conventional D-S fusion usually assumes that evidence sources have already been transformed into structured mass functions.
This limits its direct applicability to retrieval-augmented QA, where evidence is retrieved as noisy unstructured text.

Neuro-symbolic reasoning aims to combine the flexibility of neural models with the interpretability and controllability of symbolic reasoning~\cite{garcez2015neural,garcez2023neurosymbolic}.
For QA, symbolic or knowledge-enhanced components can provide structured constraints or intermediate reasoning states, while LLMs contribute semantic understanding and language-based inference.
Recent LLM-based reasoning methods also exploit chain-of-thought prompting, self-consistency, and tool-augmented reasoning~\cite{wei2022chain,wang2022self,shinn2023reflexion}.
Despite these advances, many methods still rely on implicit aggregation of evidence and rarely expose explicit reliability and uncertainty signals that can be inspected or used in final decision arbitration.
BELIEF bridges these gaps by deriving reliability-weighted basic probability assignments from structured biomedical evidence and coupling symbolic evidence fusion with LLM semantic reasoning over the same evidence representation.

Taken together, these lines of work leave a gap between unstructured retrieval and explicit uncertainty-aware decision making, which motivates BELIEF's use of structured evidence objects, reliability weighted mass assignments, and arbitration based on belief and residual uncertainty signals.

\section{Problem Definition}
\label{sec:problem_def}

Biomedical question answering (QA) aims to derive reliable decisions from heterogeneous biomedical evidence.
In this paper, we focus on closed-set biomedical QA, where each question is associated with a finite candidate answer space.
Formally, given a natural language question $q$, a retrieved evidence set $E=\{e_i\}_{i=1}^{M}$, and a task-specific answer space
\begin{equation}
\Theta = \{h_1,h_2,\dots,h_L\},
\label{eq:fod}
\end{equation}
the goal is to predict an answer $y^* \in \Theta$.
For PubMedQA, $\Theta$ corresponds to $\{\text{yes},\text{no},\text{maybe}\}$, while for multiple-choice benchmarks such as MedQA and MedMCQA, $\Theta$ corresponds to the candidate option set. Thus, although the datasets differ in source and answer format, they share a common decision structure in which the answer is selected from a finite set of mutually exclusive hypotheses.

The closed-set formulation is important because symbolic evidence fusion requires a finite frame of discernment.
Open-ended answer generation is beyond the scope of this work and is discussed as a limitation.
Under this formulation, each retrieved evidence item may provide uncertain or partial support for candidate hypotheses, for efficient symbolic fusion, BELIEF approximates this support through a primary supported hypothesis and residual uncertainty.
The objective is therefore not only to select the most likely answer, but also to model how evidence reliability, relevance, support strength, and residual uncertainty affect the final decision.

However, biomedical evidence is inherently heterogeneous and uncertain.
Each evidence item is typically represented as unstructured biomedical text, implicitly containing clinical attributes such as population, intervention, outcome, and study characteristics.
Different evidence sources may also vary substantially in reliability and may provide incomplete, partial, or differently oriented support toward candidate hypotheses.
Therefore, treating $E$ as flat textual context and relying solely on implicit neural aggregation~\cite{lewis2020retrieval,asai2023self} makes it difficult to explicitly model evidence reliability, residual uncertainty, and multi-evidence support for decision making.

To enable structured evidence reasoning, each raw evidence item $e_i$ is first transformed into a structured evidence object
\begin{equation}
\tilde{e}_i = \phi(e_i,q),
\end{equation}
where $\tilde{e}_i$ encodes clinically meaningful attributes and textual semantic content, such as PICO-style elements, study-related information, core biomedical concepts, and evidence snippets.
The structured evidence set is denoted as $\tilde{E}=\{\tilde{e}_i\}_{i=1}^{M}$.
This structured representation provides a shared evidence basis for both symbolic evidence fusion and neural semantic reasoning.

Hypothesis-related support information is then derived through an evidence annotation function conditioned on the question and the finite answer space:
\begin{equation}
a_i = \Phi(\tilde{e}_i,q,\Theta),
\end{equation}
where $a_i$ denotes the annotation signals associated with evidence item $i$, including evidence quality, relevance, support strength, and hypothesis association.
These annotation signals connect the structured evidence representation with the candidate hypotheses in $\Theta$.

Based on $a_i$, BELIEF derives a basic probability assignment (BPA)
\begin{equation}
m_i = \psi(a_i,\Theta),
\end{equation}
where $m_i(A)$ denotes the belief mass assigned by evidence item $i$ to a subset $A\subseteq\Theta$.
Multiple BPAs are then fused into an aggregated mass function
\begin{equation}
m^{*} = \bigoplus_{i=1}^{M} m_i,
\end{equation}
where $\oplus$ denotes the D-S evidence combination operator.
The fused mass function provides symbolic reasoning signals, including hypothesis-level belief scores and residual uncertainty.

In parallel, neural semantic reasoning operates on the same structured evidence set $\tilde{E}$ to capture contextual biomedical knowledge and implicit semantic relationships.
The final answer is produced by an arbitration function that integrates symbolic fusion signals and neural semantic reasoning outputs:
\begin{equation}
y^* = \mathcal{A}(\mathcal{I}_{\mathrm{arb}}),
\label{eq:arb}
\end{equation}
where $\mathcal{I}_{\mathrm{arb}}$ denotes the structured arbitration input package containing the symbolic D-S output, uncertainty signals, evidence reliability score, and neural LLM reasoning output.

\section{Methodology}
\label{sec:methodology}

\begin{figure*}[t]
\centering
\includegraphics[width=1.0\textwidth]{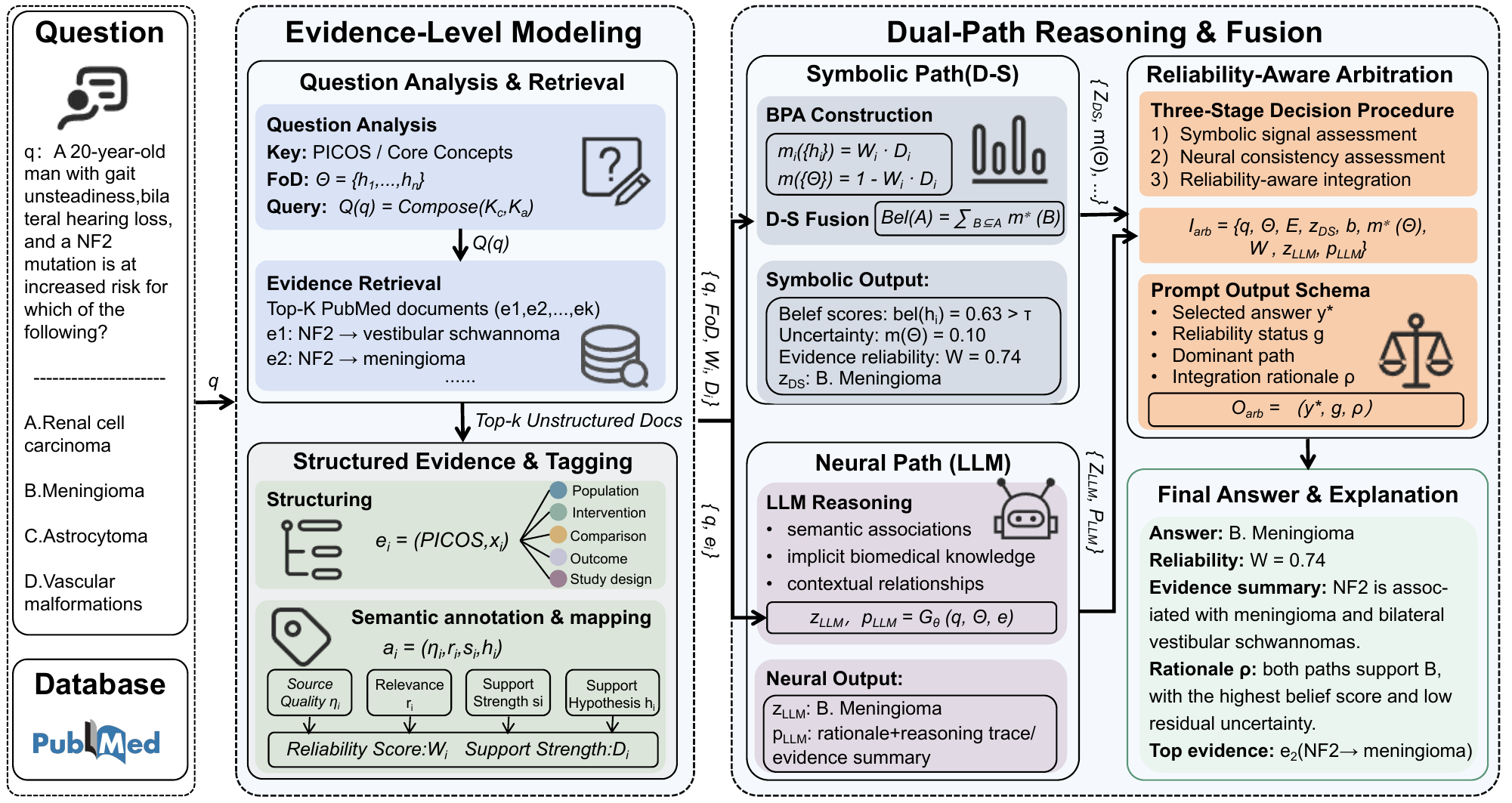}
\caption{Overview of BELIEF. Retrieved biomedical literature is converted into
structured evidence objects, processed by symbolic D-S fusion and neural semantic
reasoning, and integrated through reliability-aware arbitration.}
\label{fig:framework}
\end{figure*}

\subsection{Overview of BELIEF}
\label{sec:overview}
As shown in Fig.~\ref{fig:framework}, BELIEF first converts retrieved biomedical literature into structured evidence objects and hypothesis-conditioned annotation signals. 
These shared evidence representations are then used by two complementary reasoning paths: a symbolic D-S path that estimates belief and residual uncertainty, and a neural LLM path that performs semantic inference. 
Finally, reliability-aware arbitration integrates both outputs to produce the final decision. 
The complete inference procedure is provided in the supplementary appendix.

\subsection{Evidence-Level Modeling}
The first component focuses on transforming heterogeneous biomedical literature into structured evidence representations that are compatible with both symbolic D-S fusion and neural LLM-based semantic reasoning.
This process consists of task-aware question analysis and retrieval, structured evidence representation, and hypothesis-conditioned evidence annotation for subsequent BPA construction.

\subsubsection{Question Analysis and Evidence Retrieval}

Given a natural language question $q$, BELIEF first performs task-aware clinical analysis following evidence-based medicine (EBM) principles.
For foreground clinical questions, we extract PICOS-style elements, including Population, Intervention, Comparison, Outcome, and Study design.
For background or factual biomedical questions, where explicit intervention structures may be absent, we instead identify a set of core biomedical concepts that capture the semantic focus of the query.
This adaptive formulation ensures that the representation of the question remains aligned with its clinical or semantic characteristics.

Based on the analyzed question, BELIEF instantiates the task-specific frame of discernment $\Theta$ defined in Section~\ref{sec:problem_def}, where each hypothesis corresponds to a candidate answer option.
The answer space is fixed for subsequent evidence annotation, symbolic D-S fusion, neural LLM reasoning, and arbitration.

To retrieve relevant evidence, BELIEF constructs structured keyword queries by distinguishing between core keywords $\mathcal{K}_c$ and auxiliary keywords $\mathcal{K}_a$, and forms the retrieval query as
\begin{equation}
Q(q) = \mathrm{Compose}(\mathcal{K}_c, \mathcal{K}_a).
\end{equation}
The top-$k$ retrieved documents constitute the initial evidence pool $E=\{e_i\}_{i=1}^{M}$.
Unless otherwise specified, we use $k=5$ in the main experiments.

\begin{table*}[t]
\centering
\caption{Structured evidence schema shared by the symbolic and neural reasoning paths.}
\label{tab:evidence_schema}
\begin{tabular}{llll}
\toprule
\textbf{Field} & \textbf{Meaning} & \textbf{Value Space} & \textbf{Used By} \\
\midrule
$x_i$ & Evidence snippet or abstract segment & Text & Neural LLM path \\
$\mathrm{PICOS}_i$ & Population, intervention, comparison, outcome, and study design attributes & Structured text or empty fields & Both paths \\
$\eta_i$ & Source quality or evidence quality & Ordinal label & Symbolic D-S path \\
$r_i$ & Relevance to the question and candidate hypotheses & Ordinal label & Symbolic D-S path \\
$s_i$ & Support strength toward the primary hypothesis & Ordinal label & Symbolic D-S path \\
$h_i$ & Primary supported hypothesis & $h_i \in \Theta$ & Symbolic D-S path \\
\bottomrule
\end{tabular}
\end{table*}

\subsubsection{Structured Evidence Representation}

Although retrieved documents provide relevant biomedical information, they remain unstructured and may contain substantial noise for direct reasoning.
To address this issue, BELIEF introduces a structured evidence representation that extracts clinically salient information while preserving textual semantic context.
Specifically, each retrieved document is transformed into
\begin{equation}
\tilde{e}_i = \phi(e_i,q) = (\mathrm{PICOS}_i, x_i),
\end{equation}
where $\mathrm{PICOS}_i$ encodes structured biomedical attributes and $x_i$ denotes the associated evidence snippet or abstract segment.
For questions where full PICOS elements are not applicable, such as background biomedical questions, BELIEF fills unavailable PICOS fields with empty values and records core biomedical concepts in the structured evidence object.

This structured representation serves as the shared evidence basis for both reasoning paths.
The textual fields $(x_i,\mathrm{PICOS}_i)$ preserve information needed by the neural LLM path, while hypothesis-conditioned annotation signals are further derived from $\tilde{e}_i$ to support reliability-aware BPA construction in the symbolic D-S path.
The complete schema of the structured evidence object and its associated annotation signals is summarized in Table~\ref{tab:evidence_schema}.

To support downstream symbolic D-S reasoning, each structured evidence item is evaluated through a semantic annotation process conditioned on the question and the finite answer space:
\begin{equation}
a_i = \Phi(\tilde{e}_i,q,\Theta),
\end{equation}
where
\begin{equation}
a_i=(\eta_i,r_i,s_i,h_i)
\end{equation}
denotes the source-quality label, relevance label, support-strength label, and primary supported hypothesis, respectively.
Here, $\eta_i$ captures the quality or reliability of the evidence source, $r_i$ measures relevance to the question and candidate hypotheses, $s_i$ measures the evidential support strength, and $h_i\in\Theta$ denotes the hypothesis primarily supported by the evidence item.

In the current implementation, $\Phi(\cdot)$ is instantiated as a fixed structured annotation procedure implemented through constrained prompting. 
For each retrieved evidence item, the annotator receives only the question $q$, the candidate answer space $\Theta$, and the structured evidence item $\tilde{e}_i$, and is required to output a JSON-style record containing four fields: source-quality label, relevance label, support-strength label, and primary supported hypothesis. 
No ground-truth answer label is exposed to the annotation process. 
The annotation schema and the subsequent annotation-to-score mapping are fixed across all datasets and backbone models, ensuring that the symbolic D-S path uses the same computable evidence interface under different experimental settings.

Based on these annotation signals, BELIEF maps each evidence item into quantitative reliability and support scores. 
Specifically, we define
\begin{equation}
R_i = \mathcal{F}_{\mathrm{rel}}(\eta_i),
\quad
r_i' = \lambda(r_i),
\quad
D_i = \mathcal{G}_{\mathrm{sup}}(s_i),
\end{equation}
where $R_i$ denotes the source-quality score, $r_i'$ denotes the numeric relevance weight, and $D_i$ denotes the support strength toward the primary hypothesis. 
The relevance-adjusted reliability score used for BPA construction is then computed as
\begin{equation}
W_i = R_i r_i'.
\end{equation}
To summarize the relevance-weighted source reliability of the retrieved evidence set for arbitration, we compute
\begin{equation}
\bar{W}=
\frac{\sum_{i=1}^{M} r_i' R_i}{\sum_{i=1}^{M} r_i' + \epsilon},
\end{equation}
where $\epsilon$ is a small numerical constant used only to avoid division by
zero.
The annotation-to-score mapping is fixed across all experiments, with the implementation-level mapping table provided in the supplementary appendix.

\subsection{Dual-Path Reasoning and Fusion}

The second component focuses on integrating symbolic D-S evidence fusion with neural LLM-based semantic reasoning.
Given the structured evidence set $\tilde{E}$ and the answer space $\Theta$, BELIEF performs reasoning along two complementary paths and then combines their outputs through reliability-aware arbitration.

\subsubsection{Symbolic D-S Reasoning Path}

Based on the annotation signals $a_i=(\eta_i,r_i,s_i,h_i)$ and the derived scores $(W_i,D_i)$, the symbolic D-S path constructs a basic probability assignment (BPA) for each evidence item:
\begin{equation}
m_i = \psi(a_i,\Theta).
\end{equation}
For an evidence item primarily supporting hypothesis $h_i \in \Theta$, the BPA is defined as
\begin{equation}
m_i(\{h_i\}) = W_iD_i,
\quad
m_i(\Theta)=1-W_iD_i.
\end{equation}

This restricted formulation assigns probability mass only to the primary supported hypothesis and the ignorance set $\Theta$, allowing the symbolic D-S path to represent explicit evidence support while preserving residual uncertainty.
When an evidence item does not provide discriminative support for a specific hypothesis, its support strength is reduced and most mass remains on $\Theta$.
This formulation also keeps the fusion process computationally efficient, which is important for retrieval-augmented biomedical QA with multiple evidence items.

The symbolic D-S path then aggregates evidence using Dempster's rule of combination.
Given two mass functions $m_a$ and $m_b$, the combined mass assigned to a non-empty subset $C \subseteq \Theta$ is computed as
\begin{equation}
(m_a \oplus m_b)(C)
=
\frac{1}{Z}
\sum_{A \cap B = C} m_a(A)m_b(B),
\end{equation}
where $Z=\sum_{A,B\subseteq\Theta: A\cap B\neq \emptyset}m_a(A)m_b(B)$ normalizes
the combined mass over non-empty intersections.
By iteratively applying this operator to $\{m_i\}_{i=1}^{M}$, BELIEF obtains the fused mass function
\begin{equation}
m^{*} = \bigoplus_{i=1}^{M} m_i.
\end{equation}

Based on the fused mass distribution, the belief function is defined as
\begin{equation}
\mathrm{Bel}(A)=\sum_{B\subseteq A}m^{*}(B),
\end{equation}
which quantifies the accumulated support assigned to hypothesis set $A$.
The belief vector over candidate hypotheses is denoted as
\begin{equation}
\mathbf{b}=
\left[
\mathrm{Bel}(\{h_1\}),
\mathrm{Bel}(\{h_2\}),
\dots,
\mathrm{Bel}(\{h_L\})
\right].
\end{equation}
Based on the belief vector, the preliminary symbolic prediction is obtained as
\begin{equation}
\hat{z}_{\mathrm{DS}}
=
\arg\max_{h_j\in\Theta}
\mathrm{Bel}(\{h_j\}).
\end{equation}

To avoid overconfident symbolic decisions under insufficient evidence, BELIEF further introduces a threshold-based abstention mechanism.
If the maximum belief score is lower than a predefined threshold $\tau$, the symbolic D-S path explicitly defers the decision:
\begin{equation}
z_{\mathrm{DS}}=
\begin{cases}
\hat{z}_{\mathrm{DS}},
& \text{if }\max_j\mathrm{Bel}(\{h_j\})\ge\tau,\\
\texttt{uncertain},
& \text{otherwise.}
\end{cases}
\end{equation}

The resulting symbolic output consists of the abstention-aware symbolic decision $z_{\mathrm{DS}}$, belief vector $\mathbf{b}$, residual uncertainty mass $m^{*}(\Theta)$, and aggregate evidence reliability $\bar{W}$.
These signals provide structured reliability and uncertainty information for subsequent arbitration.

\subsubsection{Neural LLM Path}

In parallel, the neural LLM path performs semantic inference over the same structured evidence.
Formally, the LLM takes the question, candidate answer space, and structured evidence set as input and produces
\begin{equation}
z_{\mathrm{LLM}},p_{\mathrm{LLM}}
=
\mathcal{G}_{\theta}(q,\Theta,\tilde{E}),
\end{equation}
where $z_{\mathrm{LLM}}$ denotes the generated prediction, and $p_{\mathrm{LLM}}$ represents the reasoning-related output produced by the LLM, including the generated rationale and any self-reported confidence cues used only for arbitration.

Compared with the symbolic D-S path, the neural LLM path is better at capturing implicit biomedical knowledge, contextual relationships, and semantic associations that are difficult to encode through fixed evidence annotations.
However, it does not explicitly quantify hypothesis-level belief, residual uncertainty, or evidence reliability.

\subsubsection{Reliability-Aware Arbitration}
\label{subsec:arbitration}

Since the symbolic D-S path and the neural LLM path may produce different predictions, BELIEF introduces a reliability-aware arbitration module to determine the final answer.
The goal of arbitration is not to simply select the output with higher confidence, but to examine whether each reasoning path is reliable under the current evidence condition.

The arbitration module receives a structured input package:
\begin{equation}
\mathcal{I}_{\mathrm{arb}} =
\{q,\Theta,\tilde{E},z_{\mathrm{DS}},\mathbf{b},m^{*}(\Theta),\bar{W},z_{\mathrm{LLM}},p_{\mathrm{LLM}}\},
\end{equation}
where $\mathbf{b}$ denotes the belief scores over candidate hypotheses, $m^{*}(\Theta)$ denotes the residual uncertainty mass, and $\bar{W}$ denotes an aggregate evidence reliability score.
The module outputs
\begin{equation}
o_{\mathrm{arb}}=(y^*,g,\rho),
\end{equation}
where $y^*$ is the final answer, $g$ indicates the reliability status of the final decision, and $\rho$ is a concise rationale explaining the integration outcome.
The structured input and output fields of the arbitration module are summarized in Table~\ref{tab:arbitration_schema}.

The arbitration process is implemented as a structured three-stage decision procedure.
First, symbolic signal assessment examines the belief vector, residual uncertainty mass, and aggregate evidence reliability to determine whether the symbolic D-S output is sufficiently reliable under the current evidence condition.
Second, neural consistency assessment checks whether the neural LLM prediction and rationale are semantically consistent with the structured evidence and the candidate hypotheses.
Third, reliability-aware integration selects or reconciles the two path outputs according to their evidence grounding, uncertainty profile, and semantic consistency.
These steps allow BELIEF to use symbolic D-S signals as explicit reliability indicators rather than as additional textual context only.

In implementation, we use a constrained arbitration prompt template that presents these signals in fixed fields and requires a schema-constrained decision output.
The output is constrained to a fixed schema containing the selected answer, final decision reliability status, dominant reasoning path, and integration rationale.

\begin{table}[t]
\centering
\caption{Structured signals used in reliability-aware arbitration.}
\label{tab:arbitration_schema}
\setlength{\tabcolsep}{4pt}
\renewcommand{\arraystretch}{1.08}
\begin{tabular}{lp{0.68\linewidth}}
\toprule
\textbf{Signal} & \textbf{Description} \\
\midrule
$z_{\mathrm{DS}}$ & Symbolic prediction, or \texttt{uncertain} when evidence is insufficient. \\
$\mathbf{b}$ & Belief scores over candidate hypotheses in $\Theta$. \\
$m^{*}(\Theta)$ & Residual uncertainty after D-S fusion. \\
$\bar{W}$ & Overall reliability of the retrieved evidence set. \\
$z_{\mathrm{LLM}}$ & Prediction generated by the neural reasoning path. \\
$p_{\mathrm{LLM}}$ & LLM reasoning trace used for consistency inspection. \\
$y^*$ & Final answer selected by the arbitration module. \\
$g$ & Reliability status of the final decision. \\
$\rho$ & Rationale explaining the integration decision. \\
\bottomrule
\end{tabular}
\end{table}

\subsection{Properties of the Symbolic D-S Path}
\label{sec:fusion_analysis}

The symbolic D-S path in BELIEF has three useful properties: validity, uncertainty awareness, and computational efficiency. 
For each structured evidence instance $\tilde{e}_i$ annotated as $a_i=(\eta_i,r_i,s_i,h_i)$, BELIEF constructs a restricted BPA over the primary supported hypothesis $h_i\in\Theta$ and the ignorance set $\Theta$:
\begin{equation}
    m_i(\{h_i\}) = W_i D_i, \quad 
    m_i(\Theta) = 1 - W_i D_i,
\end{equation}
where $W_i \in [0,1]$ denotes evidence reliability and $D_i \in [0,1]$ denotes support strength. 
Since $W_iD_i\in[0,1]$, the constructed mass function satisfies $m_i(\emptyset)=0$, non-negativity, and normalization:
\begin{equation}
    \sum_{A \subseteq \Theta} m_i(A)
    = m_i(\{h_i\}) + m_i(\Theta)
    = 1.
\end{equation}
All other subsets receive zero mass, making the restricted BPA compatible with standard D-S evidence fusion.

This formulation is also suitable for biomedical QA because retrieved evidence often provides partial rather than conclusive support for a candidate answer. 
By preserving $m_i(\Theta)$ and the fused residual uncertainty $m^{*}(\Theta)$, the symbolic path explicitly represents insufficient or weakly discriminative evidence instead of forcing every evidence item to make a deterministic decision. 
When uncertainty is high or reliability is low, the arbitration module can reduce reliance on the symbolic output and favor the neural LLM path only when its rationale remains semantically consistent with the structured evidence.

Finally, the restricted BPA design keeps symbolic fusion lightweight. 
Standard D-S fusion over the full power set may require up to $2^{|\Theta|}$ focal elements, whereas BELIEF only maintains singleton hypotheses and the ignorance set, resulting in at most $|\Theta|+1$ active focal elements during iterative fusion. 
Therefore, the symbolic path avoids full power-set expansion and provides efficient reliability and uncertainty modeling for retrieval-augmented biomedical QA.

\section{Experiments}
\label{sec:experiments}

\subsection{Datasets}

We evaluate the proposed framework on three widely used biomedical question answering benchmarks, namely PubMedQA~\cite{jin2019pubmedqa}, MedQA~\cite{jin2021disease}, and MedMCQA~\cite{pal2022medmcqa}. 
PubMedQA is a literature-grounded yes/no/maybe task based on biomedical abstracts, and is suitable for evaluating uncertainty-aware decision-making under incomplete or ambiguous evidence. 
MedQA is derived from professional medical examinations and focuses on clinical reasoning and structured multiple-choice answer selection. 
MedMCQA is a large-scale multiple-choice benchmark covering diverse biomedical and clinical topics, providing a broader testbed for evaluating robustness and generalization. 
In our experiments, all compared methods are evaluated on the same sampled test
instances: 500 from PubMedQA, 500 from MedQA, and 1,000 from MedMCQA. Samples are randomly selected from the official test or evaluation split with a fixed random seed, for PubMedQA, we preserve the original label distribution during sampling.

\subsection{Implementation and Evaluation Metrics}

For all retrieval-based methods, we uniformly query PubMed and retain the top-$k$ retrieved documents as external evidence for each question. 
Unless otherwise specified, we set $k=5$ and use the same retrieval pipeline and evidence pool across all retrieval-based baselines. 
All experiments are conducted in a zero-shot setting with a 4096-token context window, low-temperature decoding, and consistent generation hyperparameters. 
For BELIEF, the symbolic D-S path uses a fixed abstention threshold $\tau=0.5$ without task-specific tuning, and sensitivity analyses of $k$ and $\tau$ are reported in Section~\ref{subsec:hyper}.

Structured evidence annotation in BELIEF is implemented with a fixed constrained prompt corresponding to $\Phi(\cdot)$ in Section~\ref{sec:methodology}. The prompt templates used across all stages of the BELIEF reasoning pipeline are fixed across all experiments, with condensed but structurally faithful versions provided in the supplementary appendix.
For each backbone setting, the same backbone is used for annotation, neural semantic reasoning, and reliability-aware arbitration, while the annotation schema and score mapping are fixed across datasets and models. 
The annotator only uses the question, candidate hypotheses, and retrieved evidence item as input, and no ground-truth answer label is exposed during annotation or arbitration. 
When an annotation field is invalid or missing, it is treated as non-informative and mapped to the lowest corresponding label, such as UNCLEAR\_BASIC, IRRELEVANT, or NONE, before score conversion.

We evaluate BELIEF on Qwen2.5-7B~\cite{qwen25}, Qwen3-8B, Qwen3-32B~\cite{qwen3}, GPT-4o-mini~\cite{openai20244o}, and GPT-OSS-20B~\cite{agarwal2025gpt}. 
All methods are evaluated on the same sampled evaluation sets to ensure fair comparison across different backbone models and QA paradigms.
Accuracy (ACC) is used as the primary metric, and Macro-F1 is reported as a complementary metric to assess class-balanced performance. Note that all results are evaluated within our specific experimental setup, with the code and datasets being publicly available.\footnote{https://github.com/ZUST-BIT/BELIEF}
\subsection{Baselines}
We compare BELIEF with representative retrieval-augmented and reasoning baselines: Direct Generation, which serves as a lower bound by relying solely on the LLM's internal knowledge, Naive RAG~\cite{lewis2020retrieval}, representing the classic retrieve-then-read paradigm, RAT~\cite{wang2024rat}, which integrates retrieval into the chain-of-thought process, Self-RAG~\cite{asai2023self}, which uses reflection tokens to assess evidence, and CRAG~\cite{yan2024corrective}, which triggers corrective retrieval when internal retrieval is unreliable. These baselines cover parametric-only answering, flat-context retrieval, iterative retrieval-reasoning, self-reflection, and corrective retrieval. Although they improve QA from different perspectives, they generally do not explicitly transform retrieved literature into structured evidence objects or model reliability-weighted support and residual uncertainty over candidate hypotheses.

To further compare against domain-specific biomedical models, we include BioMistral-7B~\cite{labrak2024biomistral}, Self-BioRAG-7B~\cite{jeong2024improving}, UltraMedical-8B~\cite{zhang2024ultramedical}, Huatuo-GPT-o1-8B~\cite{chen-etal-2025-towards-medical}, Llama-3-Meditron-8B~\cite{sallinen2025llama}, and MEDITRON-70B~\cite{chen2023meditron}.
These models provide a complementary comparison axis in terms of biomedical pretraining, instruction tuning, and domain alignment. 
We evaluate these models on the same sampled instances using the same answer-selection protocol when executable checkpoints are available, this comparison is intended to provide a domain-specific reference rather than to claim a fully controlled training-level comparison.

\begin{table*}[t]
\centering
\caption{Results of general LLMs under different QA paradigms. Best results are shown in bold, second-best results are underlined.}
\renewcommand{\arraystretch}{1.00}
\label{tab:general_llm_results}
\begin{tabular*}{0.96\textwidth}{@{\extracolsep{\fill}}llcccccc}
\toprule
\textbf{Backbone} & \textbf{Method} & \multicolumn{2}{c}{\textbf{PubMedQA}} & \multicolumn{2}{c}{\textbf{MedQA}} & \multicolumn{2}{c}{\textbf{MedMCQA}} \\
\cmidrule(lr){3-4} \cmidrule(lr){5-6} \cmidrule(lr){7-8}
& & \textbf{ACC} & \textbf{Macro-F1} & \textbf{ACC} & \textbf{Macro-F1} & \textbf{ACC} & \textbf{Macro-F1} \\
\midrule
\multirow{6}{*}{Qwen2.5-7B}
& Direct Generation & \underline{0.508} & 0.415 & 0.574 & 0.562 & \underline{0.563} & 0.552 \\
& Naive RAG         & 0.488 & \underline{0.421} & 0.596 & 0.586 & 0.519 & 0.512 \\
& RAT               & 0.344 & 0.328 & 0.638 & \underline{0.634} & 0.552 & 0.544 \\
& Self-RAG          & 0.476 & 0.379 & \underline{0.642} & \underline{0.634} & 0.551 & \underline{0.549} \\
& CRAG              & 0.468 & 0.390 & 0.592 & 0.585 & 0.551 & 0.541 \\
& \textbf{BELIEF}   & \textbf{0.632} & \textbf{0.447} & \textbf{0.660} & \textbf{0.645} & \textbf{0.622} & \textbf{0.612} \\
\midrule
\multirow{6}{*}{Qwen3-8B}
& Direct Generation & 0.376 & 0.348 & 0.628 & 0.626 & 0.571 & 0.570 \\
& Naive RAG         & 0.396 & 0.331 & 0.664 & 0.661 & 0.539 & 0.537 \\
& RAT               & 0.378 & 0.367 & 0.508 & 0.485 & 0.519 & 0.501 \\
& Self-RAG          & \underline{0.464} & \underline{0.397} & \textbf{0.700} & \textbf{0.695} & \underline{0.598} & \underline{0.596} \\
& CRAG              & 0.338 & 0.322 & 0.638 & 0.631 & 0.582 & 0.574 \\
& \textbf{BELIEF}   & \textbf{0.570} & \textbf{0.498} & \underline{0.672} & \underline{0.666} & \textbf{0.626} & \textbf{0.616} \\
\midrule
\multirow{6}{*}{Qwen3-32B}
& Direct Generation & \underline{0.630} & 0.526 & 0.742 & 0.738 & 0.663 & 0.660 \\
& Naive RAG         & 0.628 & \underline{0.534} & 0.764 & 0.758 & 0.665 & 0.661 \\
& RAT               & 0.556 & 0.482 & \underline{0.804} & \underline{0.803} & 0.648 & 0.645 \\
& Self-RAG          & 0.430 & 0.422 & 0.794 & 0.790 & 0.661 & 0.657 \\
& CRAG              & 0.596 & 0.501 & 0.748 & 0.743 & \underline{0.671} & \underline{0.668} \\
& \textbf{BELIEF}   & \textbf{0.652} & \textbf{0.552} & \textbf{0.812} & \textbf{0.808} & \textbf{0.674} & \textbf{0.672} \\
\midrule
\multirow{6}{*}{GPT-4o-mini}
& Direct Generation & 0.550 & 0.489 & 0.762 & 0.778 & 0.619 & 0.615 \\
& Naive RAG         & \underline{0.666} & \textbf{0.569} & 0.750 & 0.735 & 0.622 & 0.617 \\
& RAT               & 0.594 & 0.495 & 0.780 & 0.778 & \underline{0.650} & \underline{0.652} \\
& Self-RAG          & 0.526 & 0.452 & \underline{0.790} & \underline{0.785} & 0.639 & 0.634 \\
& CRAG              & 0.532 & 0.452 & 0.768 & 0.760 & 0.624 & 0.618 \\
& \textbf{BELIEF}   & \textbf{0.670} & \underline{0.538} & \textbf{0.794} & \textbf{0.789} & \textbf{0.659} & \textbf{0.655} \\
\midrule
\multirow{6}{*}{GPT-OSS-20B}
& Direct Generation & \underline{0.640} & \underline{0.614} & 0.846 & 0.842 & 0.675 & \textbf{0.689} \\
& Naive RAG         & 0.636 & 0.604 & 0.828 & 0.840 & 0.669 & 0.680 \\
& RAT               & 0.560 & 0.502 & 0.830 & 0.827 & 0.678 & 0.677 \\
& Self-RAG          & 0.538 & 0.494 & \underline{0.854} & \textbf{0.855} & 0.672 & 0.678 \\
& CRAG              & 0.630 & 0.553 & 0.846 & 0.842 & \underline{0.683} & \underline{0.681} \\
& \textbf{BELIEF}   & \textbf{0.678} & \textbf{0.642} & \textbf{0.862} & \underline{0.846} & \textbf{0.697} & 0.678 \\
\bottomrule
\end{tabular*}
\end{table*}

\begin{table}[t]
\centering
\caption{Contextual comparison with domain-specific biomedical models. All results
report accuracy (ACC).}
\renewcommand{\arraystretch}{1.00}
\label{tab:biomedical_models}
\begin{tabular}{lccc}
\toprule
\textbf{Model} & \textbf{PubMedQA} & \textbf{MedQA} & \textbf{MedMCQA} \\
\midrule
BioMistral-7B    & 0.600 & 0.391 & 0.347 \\
Self-BioRAG-7B   & 0.546 & 0.436 & 0.421 \\
UltraMedical-8B  & 0.770 & 0.727 & 0.626 \\
Huatuo-GPT-o1-8B & \underline{0.775} & 0.666 & 0.583 \\
Llama-3-Meditron-8B     & 0.768 & 0.599 & 0.482 \\
MEDITRON-70B     & \textbf{0.800} & 0.654 & \underline{0.651} \\
\midrule
\textbf{BELIEF (Qwen3-8B)} & 0.570 & \underline{0.672} & 0.626 \\
\textbf{BELIEF (Qwen3-32B)} & 0.652 & \textbf{0.812} & \textbf{0.674} \\
\bottomrule
\end{tabular}
\end{table}

\subsection{Main Results}
\label{subsec:main_results}

The comprehensive empirical results are summarized in Table~\ref{tab:general_llm_results} and Table~\ref{tab:biomedical_models}. 
Overall, BELIEF achieves the best result in 25 of 30 backbone--dataset--metric settings and improves accuracy over Naive RAG in all 15 backbone--dataset comparisons.
Compared with Direct Generation and Naive RAG baselines, BELIEF benefits from explicitly structuring retrieved biomedical evidence and modeling reliability-weighted support and residual uncertainty, rather than treating retrieved documents as undifferentiated textual context.

The gains are most evident on smaller and medium-sized backbones, where explicit evidence modeling can compensate for the limited capacity of the backbone to implicitly aggregate heterogeneous evidence. 
For example, on Qwen2.5-7B, BELIEF improves accuracy over Naive RAG on all three datasets, with gains supported by paired bootstrap resampling, and obtains the highest Macro-F1 scores among the compared methods in this backbone setting. 
On Qwen3-8B, BELIEF remains strongest on PubMedQA and MedMCQA, while Self-RAG obtains better results on MedQA. 
For larger-capacity backbones such as Qwen3-32B, GPT-4o-mini, and GPT-OSS-20B, BELIEF often achieves the best accuracy, while the Macro-F1 results are more metric- and dataset-dependent. 
In particular, some baselines obtain higher Macro-F1 in specific settings, suggesting that the proposed arbitration mechanism improves aggregate answer correctness in many cases but does not uniformly optimize class-balanced behavior across all datasets and backbones.

To further examine the robustness of these accuracy improvements, we conduct paired bootstrap resampling over test instances using 5,000 bootstrap resamples.
For each backbone--dataset setting, we compute the paired accuracy difference
between BELIEF and each baseline and regard a gain as supported when the 95\%
bootstrap confidence interval excludes zero.
Under this criterion, BELIEF shows supported accuracy gains over Naive RAG in 8 of
15 settings and over Self-RAG in 7 of 15 settings.
The remaining cases mainly correspond to small-margin comparisons, indicating that these results should be interpreted as competitive trends rather than conclusive improvements.

\begin{figure*}[t]
\centering
\includegraphics[width=\textwidth]{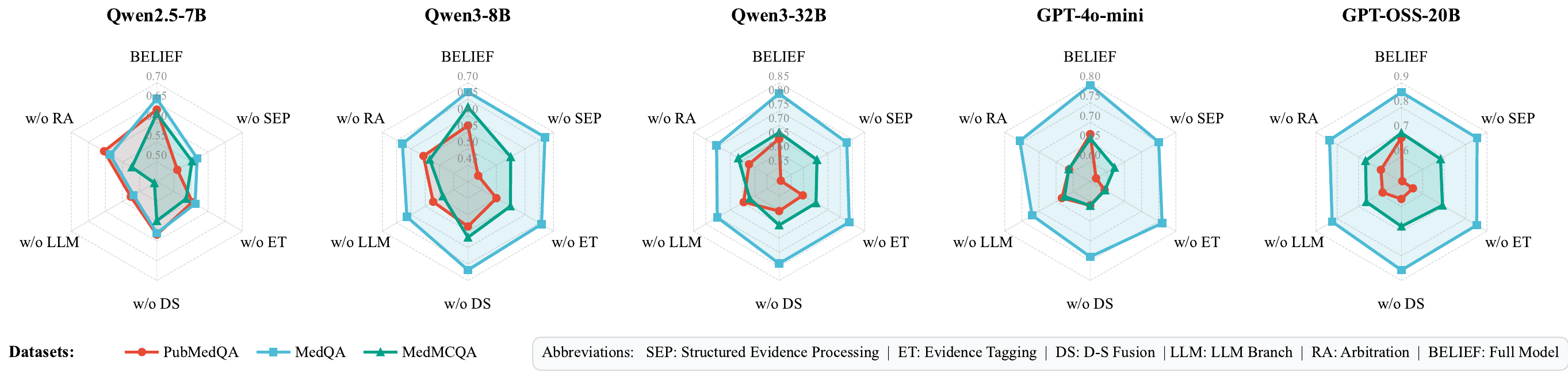}
\caption{Ablation results of BELIEF across different backbone models. Each radar plot corresponds to one backbone model, each vertex denotes the full model or one ablated variant, and each curve represents one dataset.}
\label{fig:ablation_radar}
\end{figure*}

The comparison with domain-specific biomedical models further suggests a complementary relationship between biomedical pretraining and structured evidence reasoning. 
Large biomedical models obtain higher PubMedQA accuracy in several cases, suggesting that specialized pretraining can directly support biomedical knowledge recall. 
In contrast, BELIEF obtains higher accuracy on MedQA and MedMCQA in this comparison, indicating that explicit evidence structuring, uncertainty-aware fusion, and dual-path arbitration can improve the utilization of retrieved evidence during decision making.
Therefore, the empirical findings should be interpreted as evidence that structured evidence modeling and uncertainty-aware arbitration improve retrieved-evidence utilization, rather than as a claim that BELIEF uniformly dominates all biomedical QA systems under every metric.

\subsection{Ablation Study}

\label{subsec:ablation}

To evaluate the contribution of each component in BELIEF, we construct five ablated variants by removing or simplifying one key module at a time: w/o Structured Evidence Processing (w/o SEP), w/o Evidence Tagging (w/o ET), w/o D-S Fusion (w/o DS), w/o LLM Semantic Reasoning Path (w/o LLM), and w/o Reliability-Aware Arbitration (w/o RA). The w/o RA variant replaces the prompt-guided arbitration module with a simple confidence-based path-selection strategy using symbolic belief scores and LLM self-reported confidence. The results are shown in Fig.~\ref{fig:ablation_radar}, where each radar plot corresponds to one backbone model and each curve denotes one dataset.

As shown in Fig.~\ref{fig:ablation_radar}, the full BELIEF model consistently outperforms its ablated variants across datasets and backbones, indicating that performance depends on the coordinated contribution of multiple components. Removing structured evidence processing notably reduces performance on PubMedQA, where accuracy drops from 0.570 to 0.436 on Qwen3-8B and from 0.678 to 0.504 on GPT-OSS-20B, suggesting that structured evidence objects help align retrieved biomedical literature with the answer space. Removing evidence tagging also weakens performance, e.g., on Qwen2.5-7B, accuracy decreases from 0.632 to 0.554 on PubMedQA, from 0.660 to 0.562 on MedQA, and from 0.622 to 0.536 on MedMCQA, confirming the importance of reliability-aware evidence signals for mass assignment.

The contribution of D-S fusion is reflected by the performance decline under w/o DS. On Qwen2.5-7B, accuracy decreases from 0.632 to 0.584 on PubMedQA and from 0.660 to 0.580 on MedQA, indicating that symbolic fusion provides useful belief and residual uncertainty signals beyond flat neural aggregation. The w/o LLM variant further shows that symbolic aggregation alone is insufficient for complex biomedical reasoning, as MedMCQA accuracy drops from 0.622 to 0.457 on Qwen2.5-7B and from 0.626 to 0.488 on Qwen3-8B. The w/o RA variant also consistently underperforms the full BELIEF model. On Qwen2.5-7B, replacing reliability-aware arbitration reduces accuracy from 0.632 to 0.604 on PubMedQA, from 0.660 to 0.586 on MedQA, and from 0.622 to 0.523 on MedMCQA, while similar trends are observed on GPT-4o-mini. Overall, PubMedQA is more sensitive to structured evidence processing and D-S fusion, whereas MedMCQA depends more heavily on semantic reasoning and reliability-aware arbitration, supporting the central design hypothesis of BELIEF.

\subsection{Complementarity Analysis of Dual Paths}
\label{subsec:complementarity}
\begin{table*}[t]
\centering
\caption{Complementarity and arbitration analysis of dual reasoning paths. Correctness categories are mutually exclusive.}
\label{tab:complementarity_arbitration}
\begin{tabular}{lccccccc}
\toprule
\textbf{Dataset} 
& \textbf{Both Correct} 
& \textbf{DS-only Correct} 
& \textbf{LLM-only Correct} 
& \textbf{Both Wrong} 
& \textbf{Agree Rate} 
& \textbf{Divergent Cases} 
& \textbf{Divergent Acc} \\
\midrule
PubMedQA & 192 & 61  & 114 & 133 & 0.496 & 252 & 0.492 \\
MedQA    & 162 & 36  & 135 & 167 & 0.506 & 247 & 0.543 \\
MedMCQA  & 408 & 125 & 156 & 311 & 0.490 & 510 & 0.687 \\
\bottomrule
\end{tabular}
\end{table*}

The symbolic D-S path and the neural LLM path exhibit distinct predictive behaviors, motivating complementary dual-path integration. To quantify this interaction, we analyze prediction agreement, mutually exclusive correctness patterns, and arbitration effectiveness under path divergence. Let $N$ denote the total number of instances, and let $z_{\mathrm{DS}}^{(i)}$, $z_{\mathrm{LLM}}^{(i)}$, and $\hat{y}^{(i)}$ denote the symbolic, neural, and final arbitration predictions, respectively. The agreement rate is defined as $\mathcal{R}_{\mathrm{agree}} = \frac{1}{N}\sum_{i=1}^{N}\mathbb{I}(z_{\mathrm{DS}}^{(i)} = z_{\mathrm{LLM}}^{(i)})$, the divergent set as $\mathcal{D} = \{i \mid z_{\mathrm{DS}}^{(i)} \neq z_{\mathrm{LLM}}^{(i)}\}$, and the arbitration accuracy on divergent cases as $\mathcal{R}_{\mathrm{resolve}} = \frac{1}{|\mathcal{D}|}\sum_{i \in \mathcal{D}}\mathbb{I}(\hat{y}^{(i)} = y^{(i)})$. Unless otherwise specified, results are reported using Qwen2.5-7B.

As summarized in Table~\ref{tab:complementarity_arbitration}, the agreement rate remains around 0.49--0.51 across datasets, indicating that the two paths capture different aspects of biomedical evidence and reasoning rather than behaving redundantly. The mutually exclusive correctness categories further reveal this complementarity. On PubMedQA, 61 instances are correctly answered only by the D-S path and 114 only by the LLM path, while on MedMCQA these numbers increase to 125 and 156, respectively. Although the LLM path generally exhibits stronger standalone reasoning ability, the symbolic D-S path still contributes unique evidence-grounded decisions. Meanwhile, the remaining both-wrong cases indicate that some errors stem from insufficient or ambiguous evidence rather than integration failure alone.

\begin{figure}[t]
\centering
\includegraphics[width=1.0\linewidth]{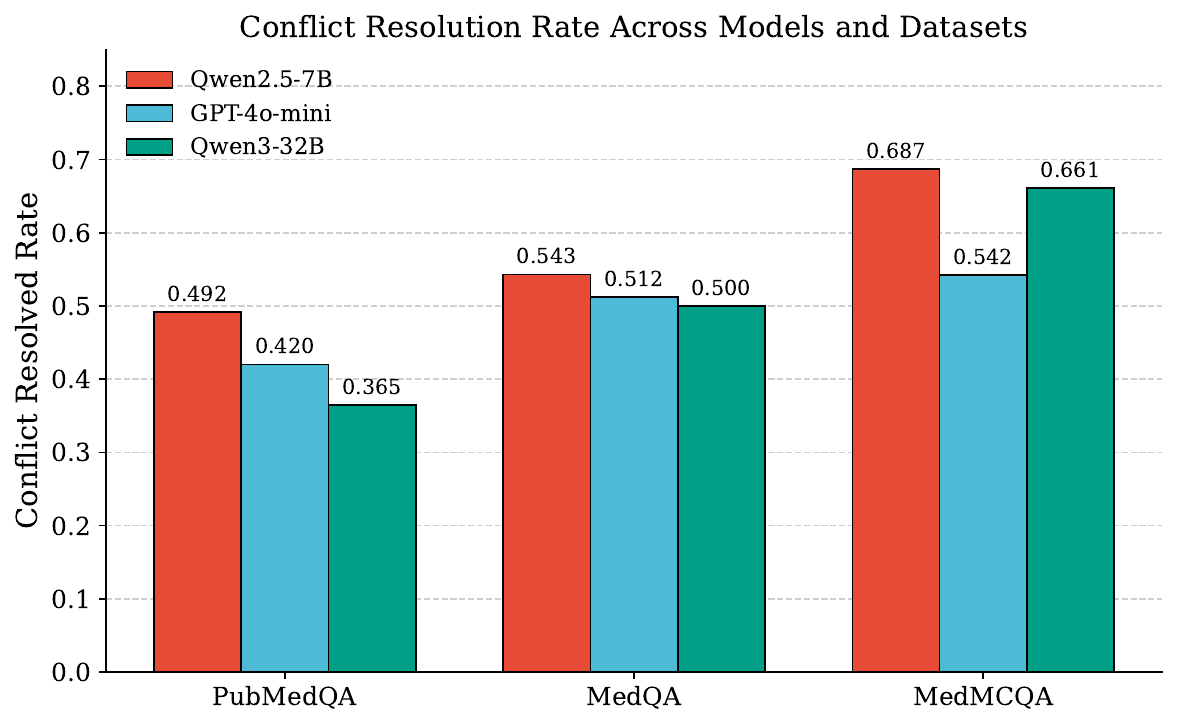}
\caption{Divergent Acc of arbitration on cases where the D-S and LLM paths produce
different predictions.}
\label{fig:path_divergence_resolution}
\end{figure}

From the arbitration perspective, a substantial number of samples fall into the divergent set (252, 247, and 510 across the three datasets), confirming that path divergence is common and final integration is not a trivial post-processing step. On these divergent cases, arbitration achieves accuracies of 0.492, 0.543, and 0.687, respectively. Compared with simple policies that always select either the D-S or LLM prediction, arbitration shows clear gains on PubMedQA and MedMCQA, where neither path is uniformly reliable, while on MedQA the advantage becomes marginal because the LLM path already dominates. As shown in Fig.~\ref{fig:path_divergence_resolution}, this behavior remains broadly consistent across backbone models. Overall, the symbolic D-S path contributes explicit evidence-grounded belief and uncertainty signals, the neural path provides flexible semantic inference, and their divergent correctness patterns motivate reliability-aware arbitration for integrating complementary reasoning signals.

\subsection{Uncertainty and Reliability Analysis}

\label{subsec:uncertainty_reliability}

To examine whether BELIEF provides practical benefits under uncertain evidence conditions, we conduct an uncertainty-stratified performance analysis. Since BELIEF explicitly estimates residual uncertainty through the D-S reasoning path, we use the fused uncertainty mass $m^{*}(\Theta)$ as the uncertainty score of each instance, where larger values indicate less sufficient or less discriminative evidence. Based on this score, test samples are divided into low ($u < 0.25$), medium ($0.25 \leq u \leq 0.55$), and high ($u > 0.55$) uncertainty groups. Both BELIEF and Naive RAG are evaluated on the same BELIEF-estimated stratified sample groups to enable direct comparison.

\begin{table}[t]
\centering
\caption{Uncertainty-stratified accuracy of BELIEF and Naive RAG.}
\label{tab:uncertainty_stratified}
\setlength{\tabcolsep}{4pt}
\renewcommand{\arraystretch}{1.05}
\begin{tabular}{llcccc}
\toprule
\textbf{Dataset} & \textbf{Level} & \textbf{\#Samples} & \textbf{BELIEF} & \textbf{Naive RAG} & $\boldsymbol{\Delta}$ \\
\midrule
\multirow{3}{*}{PubMedQA}
& Low    & 160 & 0.636 & 0.551 & +8.5\% \\
& Medium & 210 & 0.557 & 0.453 & +10.4\% \\
& High   & 130 & 0.517 & 0.382 & +13.5\% \\
\midrule
\multirow{3}{*}{MedQA}
& Low    & 180 & 0.680 & 0.633 & +4.7\% \\
& Medium & 205 & 0.671 & 0.615 & +5.6\% \\
& High   & 115 & 0.616 & 0.548 & +6.8\% \\
\midrule
\multirow{3}{*}{MedMCQA}
& Low    & 350 & 0.756 & 0.725 & +3.1\% \\
& Medium & 410 & 0.500 & 0.460 & +4.0\% \\
& High   & 240 & 0.489 & 0.444 & +4.5\% \\
\bottomrule
\end{tabular}
\end{table}

Table~\ref{tab:uncertainty_stratified} reports the accuracy of BELIEF and Naive RAG under different uncertainty levels. BELIEF consistently outperforms Naive RAG across all groups and datasets, while the performance gap generally increases as uncertainty becomes higher. For example, on MedQA, the improvement increases from 4.7 percentage points in the low-uncertainty group to 6.8 points in the high-uncertainty group, while on PubMedQA the gain increases from 8.5 to 13.5 points. This trend is particularly meaningful for PubMedQA, where biomedical literature often contains incomplete or inconclusive evidence, suggesting that explicit uncertainty estimation and reliability-aware arbitration become more beneficial when evidence is less decisive.

The stratified results also show that higher residual uncertainty generally corresponds to more difficult decision conditions. For both methods, accuracy decreases from low- to high-uncertainty groups, indicating that $m^{*}(\Theta)$ serves as a useful diagnostic signal for identifying less decisive evidence scenarios. However, BELIEF typically exhibits a smaller performance degradation. On PubMedQA, for instance, BELIEF decreases from 0.636 to 0.517 from low to high uncertainty, whereas Naive RAG drops from 0.551 to 0.382. Overall, these results support the uncertainty-aware and reliability-aware design of BELIEF. Rather than only improving average accuracy, BELIEF shows larger advantages under higher-uncertainty conditions, indicating that the uncertainty signal produced by the symbolic D-S path is useful for diagnosing harder biomedical QA scenarios.

\subsection{Hyperparameter Analysis}
\label{subsec:hyper}

\begin{figure}[t]
\centering
\includegraphics[width=1.0\linewidth]{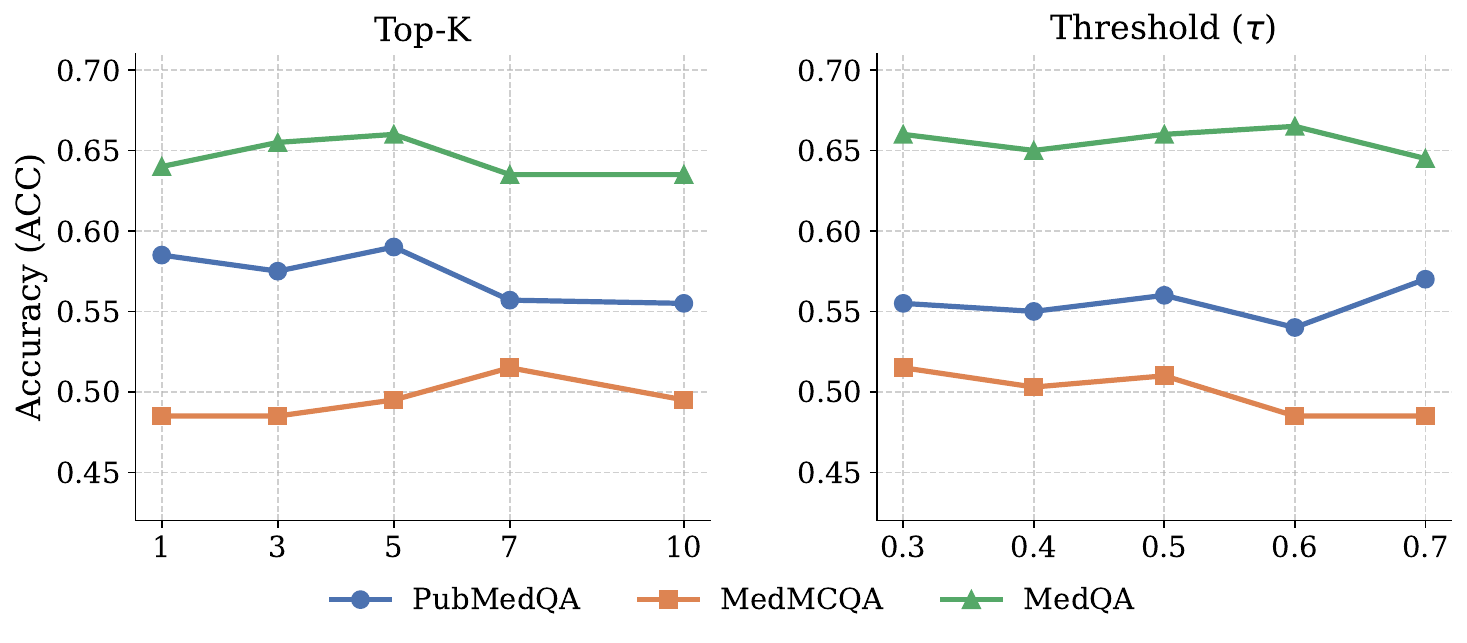}
\caption{Sensitivity analysis of BELIEF with respect to retrieval depth $k$ and abstention threshold $\tau$.}
\label{fig:hyper_analysis}
\end{figure}

We analyze the sensitivity of BELIEF to two key hyperparameters, including the retrieval depth $k$ and the abstention threshold $\tau$. 
To improve efficiency while preserving representative trends, these analyses are conducted on a 300-sample diagnostic subset drawn from each main evaluation set using Qwen2.5-7B as the backbone.
The results are summarized in Fig.~\ref{fig:hyper_analysis}, where the left panel reports the effect of retrieval depth and the right panel reports the effect of the abstention threshold.

As shown in the left panel of Fig.~\ref{fig:hyper_analysis}, the retrieval depth $k$ has a noticeable but non-monotonic effect on model performance. 
On PubMedQA and MedQA, accuracy improves as $k$ increases from 1 to 5, indicating that incorporating a moderate amount of retrieved evidence can provide more complete support for downstream reasoning. 
However, further increasing $k$ does not lead to consistent gains and may even degrade performance, suggesting that excessive retrieval can introduce noisy, weakly relevant, or redundant evidence that interferes with evidence aggregation and semantic reasoning. 
On MedMCQA, the best performance appears around $k=7$, but the improvement over $k=5$ is relatively limited. 
These results indicate that moderate retrieval depths provide a more favorable balance between evidence coverage and evidence quality. 
Therefore, we adopt $k=5$ as the unified retrieval depth in all experiments, rather than performing dataset-specific tuning.

The right panel of Fig.~\ref{fig:hyper_analysis} further evaluates the effect of the abstention threshold $\tau$ used in the symbolic reasoning path. 
Overall, BELIEF remains relatively stable across a broad range of threshold values, indicating that its performance is not overly sensitive to the precise choice of $\tau$. 
A lower threshold allows the symbolic path to participate more frequently, but may introduce less reliable symbolic decisions when the fused belief is insufficiently strong. 
In contrast, an excessively high threshold suppresses the contribution of the symbolic path and increases reliance on the neural reasoning branch, which weakens the role of explicit uncertainty-aware evidence fusion. 
Across the three datasets, moderate threshold values show more stable behavior and avoid excessive dependence on either symbolic aggregation or neural reasoning alone. 
Based on these observations, we use a fixed threshold $\tau=0.5$ for all experiments.

\subsection{Practical Cost Analysis}

To characterize the practical inference overhead of different reasoning frameworks, we conduct a cost analysis on a representative subset of PubMedQA samples using GPT-4o-mini as the backbone model. All methods are evaluated under the same inference environment and retrieval configuration to ensure fair comparison. The results are summarized in Table~\ref{tab:cost_analysis}.

In Table~\ref{tab:cost_analysis}, \textit{Avg Time} denotes the average inference time per question, while \textit{P95 Latency} represents the 95th percentile latency, meaning that 95\% of samples finish within the reported time threshold. \textit{Relative Token Cost} measures the average token consumption normalized by Naive RAG and reflects the relative computational overhead introduced by different reasoning strategies. Latency includes retrieval, generation, and method-specific reasoning steps, whereas token cost only counts language-model token usage.

Methods with more complex reasoning procedures generally incur higher computational overhead. BELIEF introduces additional cost because of structured evidence processing, dual-path reasoning, and reliability aware arbitration, its higher token consumption mainly comes from explicitly modeling evidence reliability, uncertainty, and cross-path consistency.

This overhead is accompanied by stronger reasoning capability, consistent with the performance trends observed in the main experiments. Compared with iterative reasoning frameworks such as RAT and CRAG, BELIEF allocates more computation to structured evidence integration rather than repeated corrective generation, making it more suitable for reliability-sensitive decision-support scenarios than latency-critical conversational applications.

\begin{table}[t]
\centering
\caption{Practical cost analysis of different reasoning frameworks.}
\label{tab:cost_analysis}
\begin{tabular}{lccc}
\toprule
Method & Avg Time (s) & P95 Latency (s) & Relative Token Cost \\
\midrule
Naive RAG & 8.41 & 9.69 & 1.00$\times$ \\
Self-RAG & 6.27 & 7.94 & 1.19$\times$ \\
CRAG & 95.64 & 116.35 & 1.18$\times$ \\
RAT & 26.02 & 30.93 & 1.40$\times$ \\
BELIEF & 64.31 & 77.98 & 5.21$\times$ \\
\bottomrule
\end{tabular}
\end{table}

\subsection{Case Study}
\label{subsec:case_study}

\begin{figure*}[t]
\centering
\includegraphics[width=1.00\linewidth]{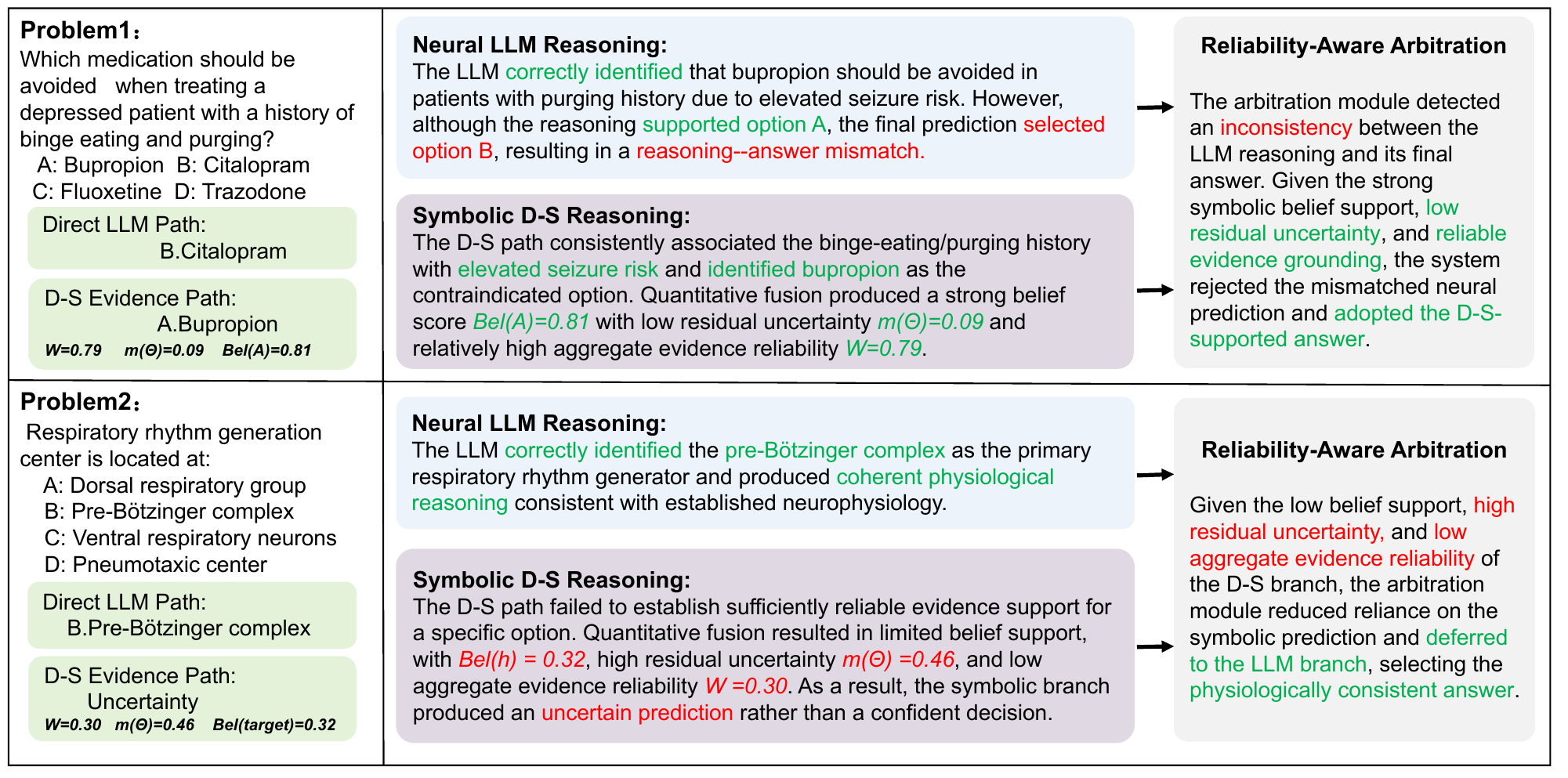}
\caption{Illustration of complementary reasoning behaviors and reliability-aware arbitration in BELIEF. The upper panel shows a reasoning--answer mismatch in the neural LLM path, which is corrected through evidence-grounded symbolic reasoning. The lower panel presents a symbolic evidence insufficiency scenario, where high uncertainty in the D-S branch causes the arbitration mechanism to defer to the neural LLM path.}
\label{fig:case_study}
\end{figure*}

This section presents representative clinical scenarios to illustrate how BELIEF performs reliability-aware integration through complementary symbolic and neural reasoning behaviors. 
As shown in Fig.~\ref{fig:case_study}, each case reports the question, candidate options, retrieved evidence summary, D-S signals, LLM output, and arbitration outcome, showing how BELIEF uses explicit belief, uncertainty, and reliability signals rather than relying solely on the generated answer.

In the first case, the LLM rationale correctly identifies that \textit{bupropion should be avoided in patients with binge-eating and purging history due to elevated seizure risk}, but the final option selected by the LLM is inconsistent with this rationale.
In contrast, the D-S path assigns high belief to the evidence-supported hypothesis, with $\mathrm{Bel}(A)=0.81$ and low residual uncertainty $m^{*}(\Theta)=0.09$. 
The arbitration module therefore identifies the LLM output as a reasoning--answer mismatch and adopts the D-S-supported answer.

In the second case, the retrieved evidence provides insufficient support for the target hypothesis, leading the D-S path to produce low belief and high uncertainty $m^{*}(\Theta)=0.46$. 
Rather than over-relying on this uncertain symbolic output, the arbitration module defers to the LLM path, whose answer is semantically coherent and supported by biomedical background knowledge. 
This case illustrates how uncertainty-aware arbitration dynamically adjusts the reliance on symbolic and neural reasoning under different evidence conditions.

\section{Limitations and Discussion}
\label{sec:limitations}

BELIEF focuses on closed-set biomedical QA where the candidate answer space forms a finite frame of discernment. It also uses a restricted BPA that assigns mass to a primary supported hypothesis and the ignorance set, which improves efficiency but does not fully model multi-hypothesis support.
Extending BELIEF to open-ended QA would require dynamic hypothesis construction or answer clustering before symbolic fusion.

The framework also depends on retrieval and annotation quality. When relevant evidence is missing or annotations are unreliable, uncertainty-aware fusion can indicate weak support but cannot by itself recover missing information. Future work may incorporate stronger retrieval models, biomedical knowledge graphs, or learned evidence-quality estimators. 

Finally, BELIEF introduces additional inference overhead due to structured evidence modeling and arbitration.
This trade-off is acceptable for reliability-sensitive analysis scenarios but may be less suitable for latency-critical applications.
Conditional arbitration and lightweight annotation models are promising directions for reducing cost.

\section{Conclusion}
\label{sec:conclusion}

We presented BELIEF, a structured evidence modeling and uncertainty-aware fusion framework for closed-set biomedical question answering.
BELIEF transforms retrieved biomedical literature into structured evidence objects, constructs D-S-theory-based mass assignments to model belief and residual uncertainty, and integrates symbolic evidence fusion with LLM-based semantic reasoning through reliability-aware arbitration.
Experiments on PubMedQA, MedQA, and MedMCQA demonstrate that BELIEF is competitive across multiple backbones and provides larger gains in higher-uncertainty groups. Diagnostic analyses further show that structured evidence modeling, uncertainty-aware symbolic fusion, and reliability-aware arbitration provide complementary benefits beyond average accuracy.
These results suggest that reliable biomedical QA can benefit not only from
stronger language models but also from explicit mechanisms for structuring,
fusing, and inspecting heterogeneous evidence under uncertainty.

\section*{Acknowledgments}
This work was supported in part by the ``Pioneer'' and ``Leading Goose`` R\&D Program of Zhejiang, China (Grant No. 2025C01115), and in part by the Graduate Research Innovation Fund of Zhejiang University of Science and Technology (Grant No. 2025yjskc07).

\bibliographystyle{IEEEtran}
\bibliography{references}

@article{brown2020language,
  title={Language models are few-shot learners},
  author={Brown, Tom and Mann, Benjamin and Ryder, Nick and Subbiah, Melanie and Kaplan, Jared D and Dhariwal, Prafulla and Neelakantan, Arvind and Shyam, Pranav and Sastry, Girish and Askell, Amanda and others},
  journal={Advances in neural information processing systems},
  volume={33},
  pages={1877--1901},
  year={2020}
}

@article{touvron2023llama,
  title={Llama: Open and efficient foundation language models},
  author={Touvron, Hugo and Lavril, Thibaut and Izacard, Gautier and Martinet, Xavier and Lachaux, Marie-Anne and Lacroix, Timoth{\'e}e and Rozi{\`e}re, Baptiste and Goyal, Naman and Hambro, Eric and Azhar, Faisal and others},
  journal={arXiv preprint arXiv:2302.13971},
  year={2023}
}

@misc{sackett1996evidence,
  title={Evidence based medicine: what it is and what it isn't},
  author={Sackett, David L and Rosenberg, William MC and Gray, JA Muir and Haynes, R Brian and Richardson, W Scott},
  journal={bmj},
  volume={312},
  number={7023},
  pages={71--72},
  year={1996},
  publisher={British Medical Journal Publishing Group}
}

@article{guyatt2011grade,
  title={GRADE guidelines: 1. Introduction—GRADE evidence profiles and summary of findings tables},
  author={Guyatt, Gordon and Oxman, Andrew D and Akl, Elie A and Kunz, Regina and Vist, Gunn and Brozek, Jan and Norris, Susan and Falck-Ytter, Yngve and Glasziou, Paul and DeBeer, Hans and others},
  journal={Journal of clinical epidemiology},
  volume={64},
  number={4},
  pages={383--394},
  year={2011},
  publisher={Elsevier}
}

@inproceedings{nye2018corpus,
  title={A corpus with multi-level annotations of patients, interventions and outcomes to support language processing for medical literature},
  author={Nye, Benjamin and Li, Junyi Jessy and Patel, Roma and Yang, Yinfei and Marshall, Iain and Nenkova, Ani and Wallace, Byron C},
  booktitle={Proceedings of the 56th Annual Meeting of the Association for Computational Linguistics (Volume 1: Long Papers)},
  pages={197--207},
  year={2018}
}

@inproceedings{lehman2019inferring,
  title={Inferring which medical treatments work from reports of clinical trials},
  author={Lehman, Eric and DeYoung, Jay and Barzilay, Regina and Wallace, Byron C},
  booktitle={Proceedings of the 2019 Conference of the North American Chapter of the Association for Computational Linguistics: Human Language Technologies, Volume 1 (Long and Short Papers)},
  pages={3705--3717},
  year={2019}
}

@article{kim2025medical,
  title={Medical hallucinations in foundation models and their impact on healthcare},
  author={Kim, Yubin and Jeong, Hyewon and Chen, Shan and Li, Shuyue Stella and Park, Chanwoo and Lu, Mingyu and Alhamoud, Kumail and Mun, Jimin and Grau, Cristina and Jung, Minseok and others},
  journal={arXiv preprint arXiv:2503.05777},
  year={2025}
}

@inproceedings{labrak2024biomistral,
  title={Biomistral: A collection of open-source pretrained large language models for medical domains},
  author={Labrak, Yanis and Bazoge, Adrien and Morin, Emmanuel and Gourraud, Pierre-Antoine and Rouvier, Mickael and Dufour, Richard},
  booktitle={Findings of the association for computational linguistics: acl 2024},
  pages={5848--5864},
  year={2024}
}

@misc{chen2023meditron,
  title={Meditron-70b: Scaling medical pretraining for large language models},
  author={Chen, Zeming and Cano, Alejandro Hern{\'a}ndez and Romanou, Angelika and Bonnet, Antoine and Matoba, Kyle and Salvi, Francesco and Pagliardini, Matteo and Fan, Simin and K{\"o}pf, Andreas and Mohtashami, Amirkeivan and others},
  journal={arXiv preprint arXiv:2311.16079},
  year={2023}
}

@inproceedings{lewis2020retrieval,
  title={Retrieval-augmented generation for knowledge-intensive NLP tasks},
  author={Lewis, Patrick and Perez, Ethan and Piktus, Aleksandra and Petroni, Fabio and Karpukhin, Vladimir and Goyal, Naman and K{\"u}ttler, Heinrich and Lewis, Mike and Yih, Wen-tau and Rockt{\"a}schel, Tim and others},
  booktitle={Advances in Neural Information Processing Systems},
  volume={33},
  pages={9459--9474},
  year={2020}
}

@article{richardson1995well,
  title={The well-built clinical question: a key to evidence-based decisions.},
  author={Richardson, W Scott and Wilson, Mark C and Nishikawa, Jim and Hayward, Robert S},
  journal={ACP journal club},
  volume={123},
  number={3},
  pages={A12--3},
  year={1995}
}

@inproceedings{garcez2015neural,
  title={Neural-Symbolic Learning and Reasoning: Contributions and Challenges.},
  author={Garcez, Artur S d'Avila and Besold, Tarek R and De Raedt, Luc and F{\"o}ldiak, Peter and Hitzler, Pascal and Icard, Thomas and K{\"u}hnberger, Kai-Uwe and Lamb, Luis C and Miikkulainen, Risto and Silver, Daniel L},
  booktitle={AAAI Spring Symposia},
  pages={18--21},
  year={2015}
}

@inproceedings{fan2024survey,
  title={A survey on rag meeting llms: Towards retrieval-augmented large language models},
  author={Fan, Wenqi and Ding, Yujuan and Ning, Liangbo and Wang, Shijie and Li, Hengyun and Yin, Dawei and Chua, Tat-Seng and Li, Qing},
  booktitle={Proceedings of the 30th ACM SIGKDD conference on knowledge discovery and data mining},
  pages={6491--6501},
  year={2024}
}

@misc{gao2024retrieval,
  title={Retrieval-Augmented Generation for Large Language Models: A Survey},
  author={Yunfan Gao and Yun Xiong and Xinyu Gao and Kangxiang Jia and Jinliu Pan and Yuxi Bi and Yi Dai and Jiawei Sun and Meng Wang and Haofen Wang},
  year={2024},
  eprint={2312.10997},
  archivePrefix={arXiv},
  primaryClass={cs.CL},
  doi={10.48550/arXiv.2312.10997},
}

@inproceedings{wang2024rat,
title={{RAT}: Retrieval Augmented Thoughts Elicit Context-Aware Reasoning and Verification in Long-Horizon Generation},
author={Zihao Wang and Anji Liu and Haowei Lin and Jiaqi Li and Xiaojian Ma and Yitao Liang},
booktitle={NeurIPS 2024 Workshop on Open-World Agents},
year={2024},
url={https://openreview.net/forum?id=5QtKMjNkjL}
}

@inproceedings{asai2023self,
  title={Self-RAG: Learning to Retrieve, Generate, and Critique through Self-Reflection},
  author={Asai, Akari and Wu, Zeqiu and Wang, Yizhong and Sil, Avirup and Hajishirzi, Hannaneh},
  booktitle={The Twelfth International Conference on Learning Representations},
  year={2024}
}

@incollection{dempster2008upper,
  title={Upper and lower probabilities induced by a multivalued mapping},
  author={Dempster, Arthur P},
  booktitle={Classic works of the Dempster-Shafer theory of belief functions},
  pages={57--72},
  year={2008},
  publisher={Springer}
}

@book{shafer1976mathematical,
  title={A Mathematical Theory of Evidence},
  author={Shafer, Glenn},
  year={1976},
  publisher={Princeton University Press}
}

@article{li2024inconsistency,
  title={Inconsistency elimination of multi-source information fusion in smart home using the Dempster-Shafer evidence theory},
  author={Li, Shijie and Xu, Hongji and Xu, Jie and Li, Xiaoman and Wang, Yang and Zeng, Jiaqi and Li, Jianjun and Li, Xinya and Li, Yiran and Ai, Wentao},
  journal={Information Processing \& Management},
  volume={61},
  number={4},
  pages={103723},
  year={2024},
  publisher={Elsevier}
}

@article{jin2022biomedical,
  title={Biomedical question answering: a survey of approaches and challenges},
  author={Jin, Qiao and Yuan, Zheng and Xiong, Guangzhi and Yu, Qianlan and Ying, Huaiyuan and Tan, Chuanqi and Chen, Mosha and Huang, Songfang and Liu, Xiaozhong and Yu, Sheng},
  journal={ACM Computing Surveys (CSUR)},
  volume={55},
  number={2},
  pages={1--36},
  year={2022},
  publisher={ACM}
}

@inproceedings{pal2022medmcqa,
  title={Medmcqa: A large-scale multi-subject multi-choice dataset for medical domain question answering},
  author={Pal, Ankit and Umapathi, Logesh Kumar and Sankarasubbu, Malaikannan},
  booktitle={Conference on health, inference, and learning},
  pages={248--260},
  year={2022},
  organization={PMLR}
}

@inproceedings{jin2019pubmedqa,
  title={PubMedQA: A Dataset for Biomedical Research Question Answering},
  author={Jin, Qiao and Dhingra, Bhuwan and Liu, Zhengping and Cohen, William and Lu, Xinghua},
  booktitle={Proceedings of the 2019 Conference on Empirical Methods in Natural Language Processing and the 9th International Joint Conference on Natural Language Processing (EMNLP-IJCNLP)},
  pages={2567--2577},
  year={2019}
}

@article{jeong2024improving,
  title={Improving medical reasoning through retrieval and self-reflection with retrieval-augmented large language models},
  author={Jeong, Minbyul and Sohn, Jiwoong and Sung, Mujeen and Kang, Jaewoo},
  journal={Bioinformatics},
  volume={40},
  number={Supplement\_1},
  pages={i119--i129},
  year={2024},
  publisher={Oxford University Press}
}

@article{zhang2024ultramedical,
  title={Ultramedical: Building specialized generalists in biomedicine},
  author={Zhang, Kaiyan and Zeng, Sihang and Hua, Ermo and Ding, Ning and Chen, Zhang-Ren and Ma, Zhiyuan and Li, Haoxin and Cui, Ganqu and Qi, Biqing and Zhu, Xuekai and others},
  journal={Advances in Neural Information Processing Systems},
  volume={37},
  pages={26045--26081},
  year={2024}
}

@inproceedings{chen-etal-2025-towards-medical,
    title={Towards Medical Complex Reasoning with {LLM}s through Medical Verifiable Problems},
    author={Junying Chen and Zhenyang Cai and Ke Ji and Xidong Wang and Wanlong Liu and Rongsheng Wang and Benyou Wang},
    booktitle={Findings of the Association for Computational Linguistics: ACL 2025},
    year={2025},
    pages={14552--14573},
}

@inproceedings{sallinen2025llama,
  title={Llama-3-meditron: An open-weight suite of medical llms based on llama-3.1},
  author={Sallinen, Alexandre and Solergibert, Antoni-Joan and Zhang, Michael and Boy{\'e}, Guillaume and Dupont-Roc, Maud and Theimer-Lienhard, Xavier and Boisson, Etienne and Bernath, Bastien and Hadhri, Hichem and Tran, Antoine and others},
  booktitle={Workshop on Large Language Models and Generative AI for Health at AAAI 2025},
  year={2025}
}

@article{qwen25,
  title={Qwen2.5 Technical Report},
  author={Yang, An and Yang, Baosong and Zhang, Beichen and Hui, Binyuan and Zheng, Bo and Yu, Bowen and Li, Chengyuan and Liu, Dayiheng and Huang, Fei and Wei, Haoran and others},
  journal={arXiv preprint arXiv:2412.15115},
  year={2024}
}

@article{qwen3,
  title={Qwen3 Technical Report},
  author={Yang, An and Li, Anfeng and Yang, Baosong and Zhang, Beichen and Hui, Binyuan and Zheng, Bo and Yu, Bowen and Gao, Chang and Huang, Chengen and Lv, Chenxu and others},
  journal={arXiv preprint arXiv:2505.09388},
  year={2025}
}

@article{agarwal2025gpt,
  title={gpt-oss-120b \& gpt-oss-20b model card},
  author={Agarwal, Sandhini and Ahmad, Lama and Ai, Jason and Altman, Sam and Applebaum, Andy and Arbus, Edwin and Arora, Rahul K and Bai, Yu and Baker, Bowen and Bao, Haiming and others},
  journal={arXiv preprint arXiv:2508.10925},
  year={2025}
}

@inproceedings{huang2006evaluation,
  title={Evaluation of PICO as a knowledge representation for clinical questions},
  author={Huang, Xiaoli and Lin, Jimmy and Demner-Fushman, Dina},
  booktitle={AMIA annual symposium proceedings},
  volume={2006},
  pages={359},
  year={2006}
}

@article{garcez2023neurosymbolic,
  title={Neurosymbolic AI: The 3rd Wave},
  author={Garcez, Artur d’Avila and Lamb, Luis C},
  journal={Artificial Intelligence Review},
  volume={56},
  number={11},
  pages={12387--12406},
  year={2023},
  publisher={Springer}
}

@article{jin2021disease,
  title={What disease does this patient have? a large-scale open domain question answering dataset from medical exams},
  author={Jin, Di and Pan, Eileen and Oufattole, Nassim and Weng, Wei-Hung and Fang, Hanyi and Szolovits, Peter},
  journal={Applied Sciences},
  volume={11},
  number={14},
  pages={6421},
  year={2021},
  publisher={MDPI}
}

@article{shinn2023reflexion,
  title={Reflexion: Language agents with verbal reinforcement learning},
  author={Shinn, Noah and Cassano, Federico and Gopinath, Ashwin and Narasimhan, Karthik and Yao, Shunyu},
  journal={Advances in neural information processing systems},
  volume={36},
  pages={8634--8652},
  year={2023}
}

@article{singhal2025toward,
  title={Toward expert-level medical question answering with large language models},
  author={Singhal, Karan and Tu, Tao and Gottweis, Juraj and Sayres, Rory and Wulczyn, Ellery and Amin, Mohamed and Hou, Le and Clark, Kevin and Pfohl, Stephen R and Cole-Lewis, Heather and others},
  journal={Nature medicine},
  volume={31},
  number={3},
  pages={943--950},
  year={2025},
  publisher={Nature Publishing Group US New York}
}

@article{huang2025survey,
  title={A survey on hallucination in large language models: Principles, taxonomy, challenges, and open questions},
  author={Huang, Lei and Yu, Weijiang and Ma, Weitao and Zhong, Weihong and Feng, Zhangyin and Wang, Haotian and Chen, Qianglong and Peng, Weihua and Feng, Xiaocheng and Qin, Bing and others},
  journal={ACM Transactions on Information Systems},
  volume={43},
  number={2},
  pages={1--55},
  year={2025},
  publisher={ACM}
}

@techreport{sentz2002combination,
title={Combination of Evidence in Dempster-Shafer Theory},
author={Sentz, Kari and Ferson, Scott},
institution={Sandia National Laboratories},
year={2002}
}

@inproceedings{wei2022chain,
  title={Chain-of-thought prompting elicits reasoning in large language models},
  author={Wei, Jason and Wang, Xuezhi and Schuurmans, Dale and Bosma, Maarten and Xia, Fei and Chi, Ed and Le, Quoc V and Zhou, Denny and others},
  booktitle={Advances in Neural Information Processing Systems},
  volume={35},
  pages={24824--24837},
  year={2022}
}

@inproceedings{wang2022self,
  title={Self-Consistency Improves Chain of Thought Reasoning in Language Models},
  author={Wang, Xuezhi and Wei, Jason and Schuurmans, Dale and Le, Quoc and Chi, Ed and Narang, Sharan and Chowdhery, Aakanksha and Zhou, Denny},
  booktitle={The Eleventh International Conference on Learning Representations},
  year={2023}
}

@article{marshall2020trialstreamer,
  title={Trialstreamer: A living, automatically updated database of clinical trial reports},
  author={Marshall, Iain J. and Nye, Benjamin and Kuiper, Jo{\"e}l and Noel-Storr, Anna and Marshall, Rachel and Maclean, Rory and Soboczenski, Frank and Nenkova, Ani and Thomas, James and Wallace, Byron C.},
  journal={Journal of the American Medical Informatics Association},
  volume={27},
  number={12},
  pages={1903--1912},
  year={2020},
  doi={10.1093/jamia/ocaa163}
}

@article{marshall2016robotreviewer,
  title={RobotReviewer: evaluation of a system for automatically assessing bias in clinical trials},
  author={Marshall, Iain J. and Kuiper, Jo{\"e}l and Wallace, Byron C.},
  journal={Journal of the American Medical Informatics Association},
  volume={23},
  number={1},
  pages={193--201},
  year={2016},
  doi={10.1093/jamia/ocv044}
}

@article{pradeep2025empowering,
  title={Empowering explainable artificial intelligence through case-based reasoning: A comprehensive exploration},
  author={Pradeep, Preeja and Caro-Mart{\'\i}nez, Marta and Wijekoon, Anjana},
  journal={IEEE Transactions on Knowledge and Data Engineering},
  year={2025},
  publisher={IEEE}
}

@article{page2021prisma,
  title={The PRISMA 2020 statement: an updated guideline for reporting systematic reviews},
  author={Page, Matthew J. and McKenzie, Joanne E. and Bossuyt, Patrick M. and Boutron, Isabelle and Hoffmann, Tammy C. and Mulrow, Cynthia D. and Shamseer, Larissa and Tetzlaff, Jennifer M. and Akl, Elie A. and Brennan, Sue E. and Chou, Roger and Glanville, Julie and Grimshaw, Jeremy M. and Hr{\'o}bjartsson, Asbj{\o}rn and Lalu, Manoj M. and Li, Tianjing and Loder, Elizabeth W. and Mayo-Wilson, Evan and McDonald, Steve and McGuinness, Luke A. and Stewart, Lesley A. and Thomas, James and Tricco, Andrea C. and Welch, Vivian A. and Whiting, Penny and Moher, David},
  journal={BMJ},
  pages={n71},
  year={2021},
  doi={10.1136/bmj.n71}
}

@misc{openai20244o,
title={GPT-4o mini: Advancing Cost-Efficient Intelligence},
author={{OpenAI}},
year={2024},
howpublished={\url{https://openai.com/index/gpt-4o-mini-advancing-cost-efficient-intelligence/}}
}

@article{yan2024corrective,
  title={Corrective Retrieval Augmented Generation},
  author={Yan, Shi-Qi and Gu, Jia-Chen and Zhu, Yun and Ling, Zhen-Hua},
  journal={arXiv preprint arXiv:2401.15884},
  year={2024},
  doi={10.48550/arXiv.2401.15884}
}

@article{shi2025final,
  title={Final: Combining first-order logic with natural logic for question answering},
  author={Shi, Jihao and Ding, Xiao and Hui, Siu Cheung and Yan, Yuxiong and Zhao, Hengwei and Liu, Ting and Qin, Bing},
  journal={IEEE Transactions on Knowledge and Data Engineering},
  year={2025},
  publisher={IEEE}
}

@article{xiao2022generalized,
  title={Generalized divergence-based decision making method with an application to pattern classification},
  author={Xiao, Fuyuan and Wen, Junhao and Pedrycz, Witold},
  journal={IEEE transactions on knowledge and data engineering},
  volume={35},
  number={7},
  pages={6941--6956},
  year={2022},
  publisher={IEEE}
}

@article{su2024knowledge,
  title={Knowledge graph neural network with spatial-aware capsule for drug-drug interaction prediction},
  author={Su, Xiaorui and Zhao, Bowei and Li, Guodong and Zhang, Jun and Hu, Pengwei and You, Zhuhong and Hu, Lun},
  journal={IEEE journal of biomedical and health informatics},
  volume={29},
  number={3},
  pages={1771--1781},
  year={2024},
  publisher={IEEE}
}

@article{liao2026enhancing,
  title={Enhancing Large Language Models Reasoning Via Multi-Path Optimization on Knowledge Graph},
  author={Liao, Jiyong and Liu, Chubo and Ding, Yan and Wang, Haotian and Tang, Zhuo and Li, Kenli and Li, Keqin},
  journal={IEEE Transactions on Knowledge and Data Engineering},
  year={2026},
  publisher={IEEE}
}

@article{li2025framework,
  title={A Framework of Knowledge Graph-Enhanced Large Language Model Based on Global Planning},
  author={Li, Yading and Song, Dandan and Tian, Yuhang and Wang, Hao and Zhou, Changzhi and Zhang, Shuhao},
  journal={IEEE Transactions on Knowledge and Data Engineering},
  volume={38},
  number={2},
  pages={736--748},
  year={2025},
  publisher={IEEE}
}

@article{xu2025large,
  title={Are large language models really good logical reasoners? a comprehensive evaluation and beyond},
  author={Xu, Fangzhi and Lin, Qika and Han, Jiawei and Zhao, Tianzhe and Liu, Jun and Cambria, Erik},
  journal={IEEE Transactions on Knowledge and Data Engineering},
  volume={37},
  number={4},
  pages={1620--1634},
  year={2025},
  publisher={IEEE}
}
%\vfill
%\newpage
%\input{material}
\end{document}